\documentclass[a4paper,british]{extarticle}
\usepackage{fullpage}
\usepackage[T1]{fontenc}
\usepackage[utf8x]{inputenc}
\usepackage{babel}
\usepackage[bottom]{footmisc}
\usepackage{float}
\usepackage{authblk}
\usepackage{mathtools}
\usepackage{amsmath}
\usepackage{amsthm}
\usepackage{amssymb}
\usepackage{graphicx}
\usepackage{algorithm2e}
\usepackage{subfig}
\usepackage{breakurl}
\usepackage{nomencl}
\usepackage{booktabs}
\usepackage{stackengine}
\usepackage{adjustbox}
\restylefloat{table}
\usepackage[
 unicode=true,
 bookmarks=true,
 bookmarksnumbered=false,
 bookmarksopen=false,
 breaklinks=true,
 pdfborder={0 0 0},
 pdfborderstyle={},
 backref=false,
 colorlinks=false
]{hyperref}
\hypersetup{
 pdftitle={Ensembling methods for countrywide short term forecasting of gas demand}
}
\usepackage[capitalise]{cleveref}

\makeatletter

\newtheorem{definition}{Definition}

\theoremstyle{plain}

\theoremstyle{plain}

\ifx\proof\undefined
\newenvironment{proof}[1][\protect\proofname]{\par
\normalfont\topsep6\p@\@plus6\p@\relax
\trivlist
\itemindent\parindent
\item[\hskip\labelsep\scshape #1]\ignorespaces
}{%
\endtrivlist\@endpefalse
}
\providecommand{\proofname}{Proof}
\fi

\@ifundefined{showcaptionsetup}{}{%
 \PassOptionsToPackage{caption=false}{subfig}}
\makeatother
\newcommand{\abs}[1]{\lvert#1\rvert}

\providecommand{\corollaryname}{Corollary}
\providecommand{\theoremname}{Theorem}
\providecommand{\keywords}[1]{\textbf{\textit{Keywords---}} #1}

\begin{document}

\title{Ensembling methods for countrywide short term forecasting of gas demand}

\author[1]{Andrea Marziali}
\author[1]{Emanuele Fabbiani}
\author[1]{Giuseppe De Nicolao}
\affil[1]{Department of Electrical, Computer and Biomedical Engineering, University of Pavia}
\maketitle

\begin{abstract}
Gas demand is made of three components: Residential, Industrial, and Thermoelectric Gas Demand. Herein, the one-day-ahead prediction of each component is studied, using Italian data as a case study. Statistical properties and relationships with temperature are discussed, as a preliminary step for an effective feature selection. Nine "base forecasters" are implemented and compared: Ridge Regression, Gaussian Processes, Nearest Neighbours, Artificial Neural Networks, Torus Model, LASSO, Elastic Net, Random Forest, and Support Vector Regression (SVR). Based on them, four ensemble predictors are crafted: simple average, weighted average, subset average, and SVR aggregation. We found that ensemble predictors perform consistently better than base ones. Moreover, our models outperformed Transmission System Operator (TSO) predictions in a two-year out-of-sample validation. Such results suggest that combining predictors may lead to significant performance improvements in gas demand forecasting.
\end{abstract} 
\hspace{10pt}


\newpage{}

\section{Introduction}
Natural gas is one of the most important energy sources in Italy: it feeds domestic and industrial heating, production processes and thermoelectic power plants. Data from SNAM Rete Gas, the Italian Transmission System Operator (TSO), show that the total Gas Demand (GD) is made of three main components: Residential Gas Demand (RGD), Industrial Gas Demand (IGD), and Thermoelectric Gas Demand (TGD). In 2018, RGD  accounted for 41.5\% of the total consumption, IGD for 25.4\% and TGD for the remaining 33.1\% \cite{snam2017report}. 

Accurate forecasts of the overall GD, as well as of its three main components, are of primary importance to energy providers, in order to improve pipe reservation and stock planning and also prevent financial penalties due to network unbalance. Moreover, GD is closely correlated with natural gas price, which is a key input for determining the optimal production plan of thermal power plants.

Several works addressed the forecasting of natural gas demand: comprehensive reviews are \cite{soldo2012forecasting} and \cite{vsebalj2017predicting}. The latter proposes a classification along four dimensions: geographical area, time horizon, method, and inputs. Herein, we are interested in country-level, one-day-ahead predictions, based on statistical learning models that leverage past gas demand, temperature and calendar features as input variables. 

With respect to prediction horizon, a distinction is made between long-term forecasting, featuring an horizon of months or years and short-term, with an horizon of one or few days. Focusing on country-wide predictions, a long-term model based on temperature was proposed in \cite{SARAK2003929} to forecast Turkish demand. The importance of the relation between weather and gas demand was also highlighted in \cite{BALDACCI2016190} and \cite{gil2004generalized}. In \cite{potovcnik2007forecasting} a statistical model was applied to forecast the long-term evolution of Slovenian demand, while different kinds of so-called "grey models" were applied in \cite{Wu2015grey} and \cite{Zeng2016china} to forecast Chinese demand. \\
In \cite{zhu2015short}, short-term forecasting of UK natural gas demand was addressed using support vector regression with false neighbours filtered. According to the authors, the method performed better than Auto-Regressive Moving Average (ARMA) models and neural networks (ANN). Azadeh et al. \cite{azadeh2010adaptive} proposed an adaptive network-based fuzzy inference system (ANFIS) to predict Iranian gas demand, which improved on classical time series methods and ANN. A more advanced model, combining wavelet transform, genetic algorithm, ANFIS and ANN was applied in \cite{panapakidis2017day} to the Greek gas distribution network.

Long-term evolution of the Italian gas demand is investigated in \cite{Bianco2014residential} and \cite{Bianco2014nonres}: macroeconomic indicators, such as gross domestic product and gas prices and climatic factors are used to build scenarios of RGD and overall GD evolution up to 2030. However, to the best of our knowledge, the daily series of Italian GD and its peculiar features have not been studied in the literature and no result about its short-term forecasting has been presented.

A previous work \cite{fabbiani2019forecasting}, focusing on Italian RGD, proposed and compared five prediction models: ridge regression, Gaussian Process (GP), nearest neighbours, Artificial Neural Networks (ANN), and torus model, concluding that ANN and GP provided the best results. Herein, we extend the analysis along three directions: first, also the prediction of IGD and TGD is addressed, thus enabling the prediction of the overall Italian GD; second, four additional base forecasters (LASSO, elastic net, random forest, and support vector regression) are considered; third and finally, the use of ensemble predictors, i.e. forecasters obtained by the suitable aggregation of base forecasts, is investigated. More precisely, based on the nine base models (the five discussed in \cite{fabbiani2019forecasting} and the four additional ones), four ensemble predictors are considered: simple average, weighted average, subset average, and support vector regression aggregation.

Ensembling, also known as blending, is known to be an effective technique to improve overall accuracy and stability, see e.g. \cite{armstrong2001principles, hyndman2018forecasting}. Recently, ensemble predictors have been proven successful in forecasting electric load \cite{Nowotarski2016combining}, whose series shows a periodic structure similar to the one of GD.

The contribution of this paper is thus threefold: first, we present and discuss the statistical properties of Italian IGD and TGD; on these data, we develop, apply, and compare nine machine-learning models; finally, we explore the use of ensemble predictors, assessing the consequent improvements.

The paper is organised as follows. In \cref{problem_statement} we present the forecasting problem and the available data, while in \cref{exploratory_analisys} we describe the most relevant features of IGD and TGD time series. After discussing feature engineering (\cref{feature_selection}), in \cref{predictive_models} the adopted models are concisely overviewed. Training and hyperparameter tuning are presented in \cref{experiments}. Results are reported in \cref{results} and some concluding remarks (\cref{conclusion}) end the paper.

\section{Problem statement} \label{problem_statement}

In this paper, the prediction of the Italian daily GD is addressed both at aggregated and disaggregated level. Apart from minor components that can be neglected, at any day $t$, the overall GD is given by the sum of Industrial Gas Demand (IGD), Thermoelectric Gas Demand (TGD) and Residential Gas Demand (RGD):
\begin{equation*}
    \mathrm{GD}_{t} = \mathrm{IGD}_{t} + \mathrm{TGD}_{t} + \mathrm{RGD}_{t}
\end{equation*}
IGD includes demand by industrial plants, while TGD only accounts for the fuel required by thermoelectric power plants. The residential component and its forecasting was studied in a companion paper \cite{fabbiani2019forecasting} and is therefore not further discussed herein. \\
Three one-day-ahead forecasting problems are considered: (i) IGD, (ii) TGD, and (iii) overall GD. The last forecast will be obtained by summing (i) and (ii) with the prediction of RGD. 

\begin{figure}[H]
    \centering
	\subfloat{
	    \includegraphics[width=.50\textwidth]{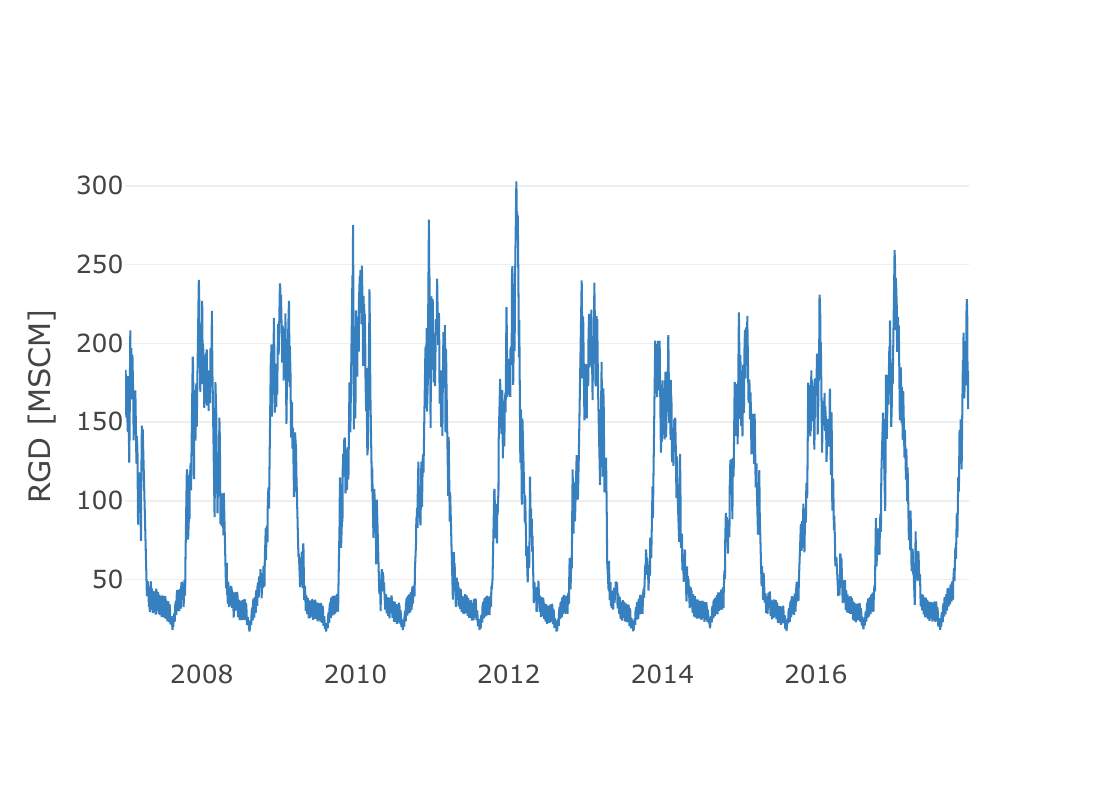}
	 }
	\subfloat{
	    \includegraphics[width=.50\textwidth]{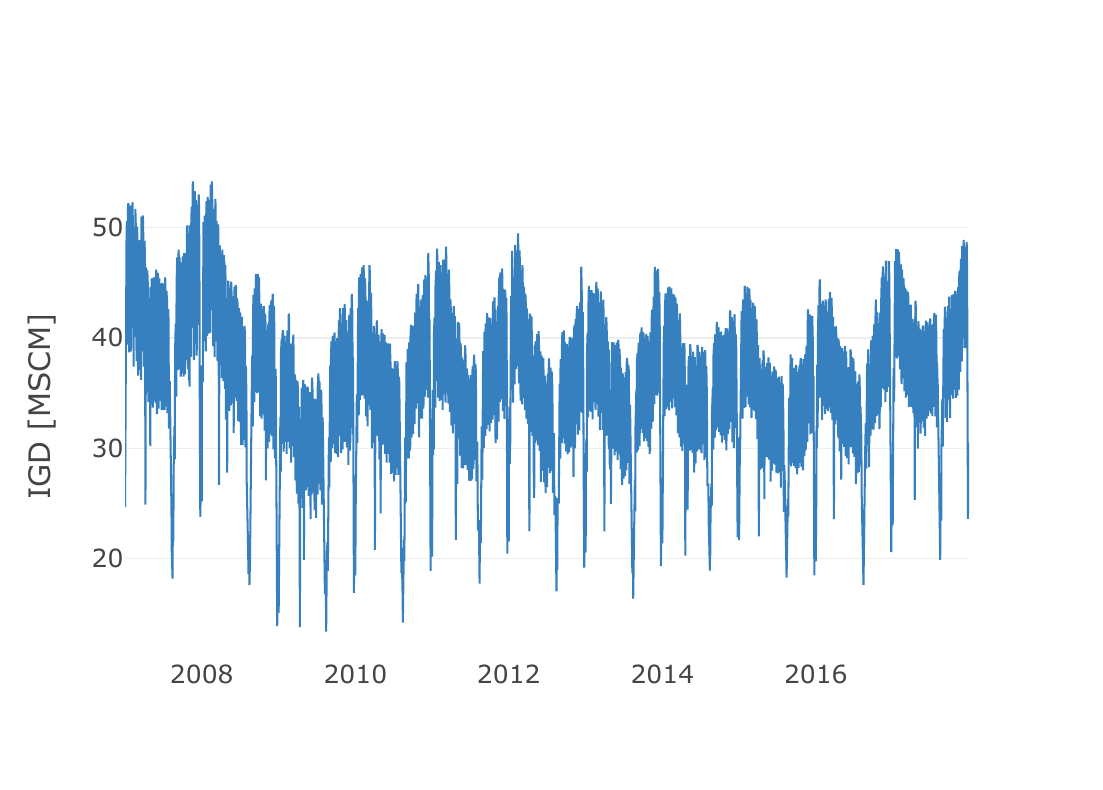}
	 } \\
	 \subfloat{
	    \includegraphics[width=.50\textwidth]{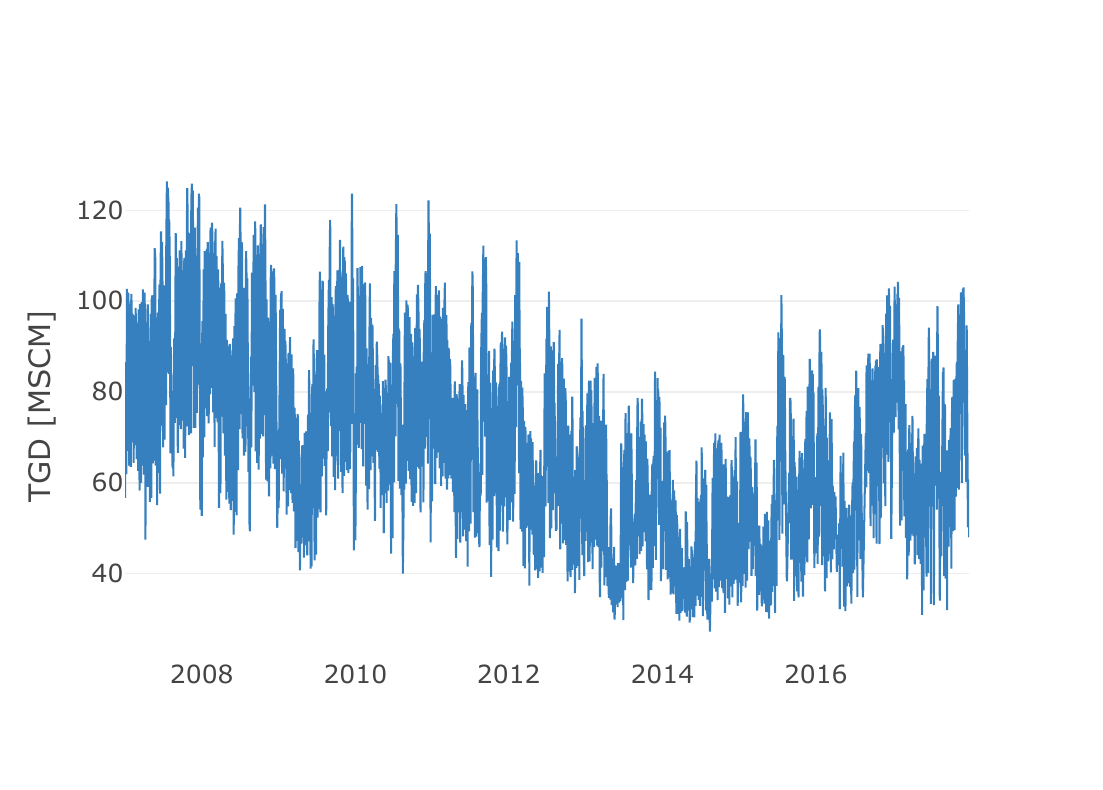}
	 }
	\subfloat{
	    \includegraphics[width=.50\textwidth]{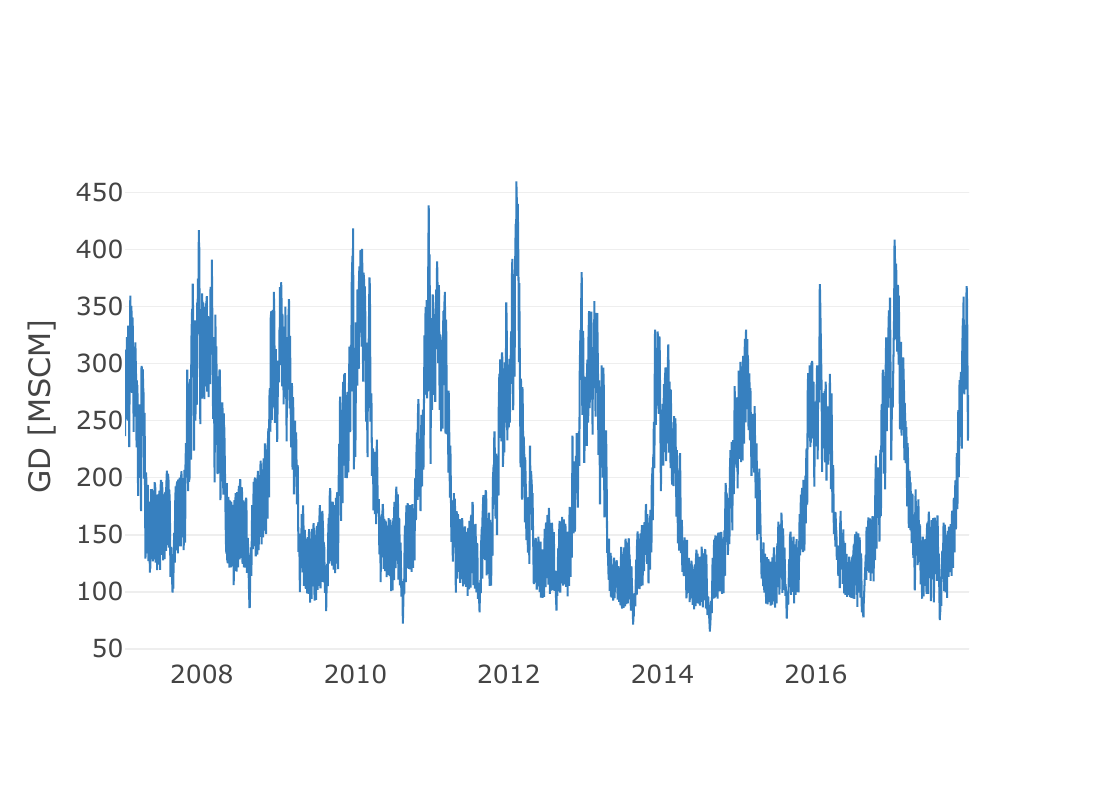}
	 }
	\caption{Top left: Italian Residential Gas Demand (RGD); top right: Italian Industrial Gas Demand (IGD); bottom left: Italian Thermoelectic Gas Demand (TGD); bottom right: overall Italian Gas Demand ($\mathrm{GD}=\mathrm{RGD}+\mathrm{IGD}+\mathrm{TGD}$).}
    \label{GD_vs_time}
\end{figure}

The datasets for both TGD and IGD are 12 year long, ranging from 2007 to 2018, and consist of 3 fields: date ($t$), forecasted average temperature in Northern Italy ($T$)\footnote{Weather forecast were provided by an Italian specialised company}, and gas demand (either IGD or TGD). Temperature in Northern Italy was considered as this region has the most rigid climate, and is thus more sensible to heating requirements. In \cref{GD_vs_time} the complete series of RGD, IGD, TGD and overall GD are displayed.

\section{Exploratory analysis} \label{exploratory_analisys}

\subsection{Industrial gas demand}

\begin{figure}[H]
    \centering			\includegraphics[width=\textwidth]{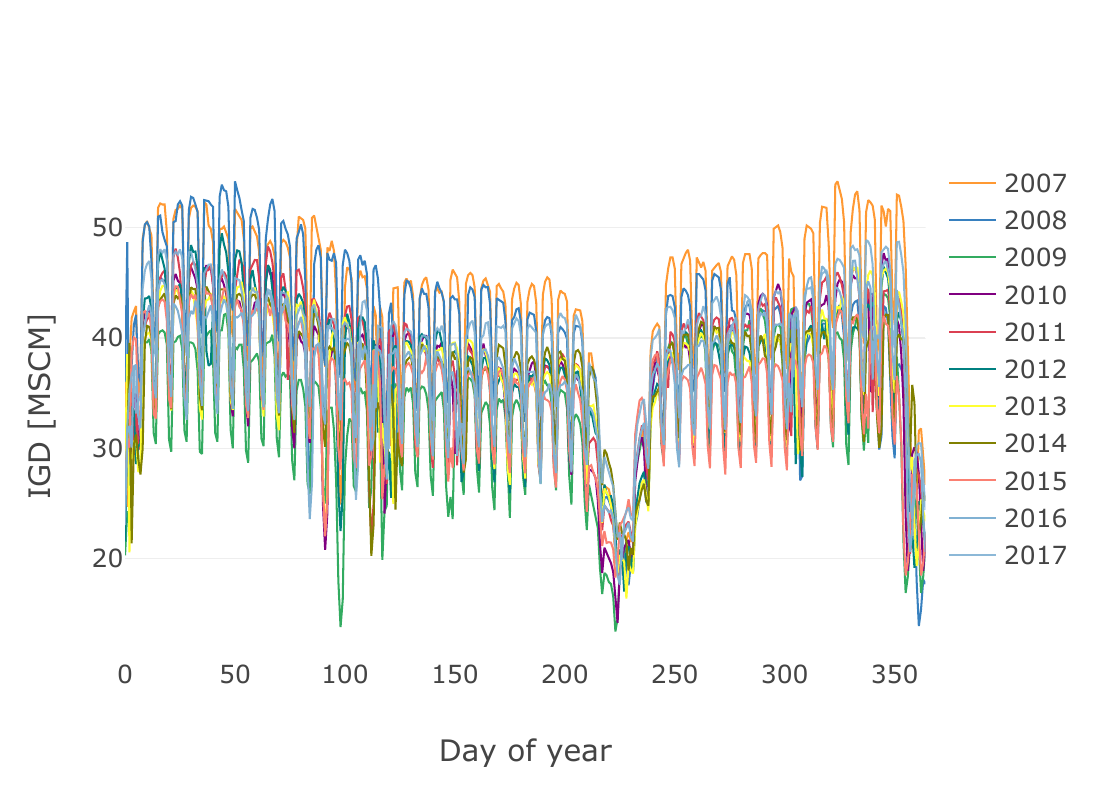}
    \caption{IGD, years 2007-2017. The time series has been shifted to align weekdays. \label{igd_superimp}}
\end{figure}

Industrial Gas Demand (IGD) does not exhibit strong trends: a significant decrease is only recorded in 2009, following the financial crisis started in the previous year. The series presents  weekly and yearly seasonal patterns. In particular, as most of the industrial facilities stop or slow down production during the weekend, IGD is lower on Saturdays and Sundays. In August and at the end of December, typical holiday periods, IGD drops to about half of its average value. Other holidays, such as Easter and the Labour Day, result in similar effects.
During the year, IGD shows a decrease from January to August and an increase from September to December, due to the use of gas for environmental heating.
All these features can be appreciated in \cref{igd_superimp}, where 11 years of IGD are superimposed, aligning weekdays to better highlight periodic behaviours.

\begin{figure}[H]
	\centering
	\subfloat{\includegraphics[width=.48\textwidth]{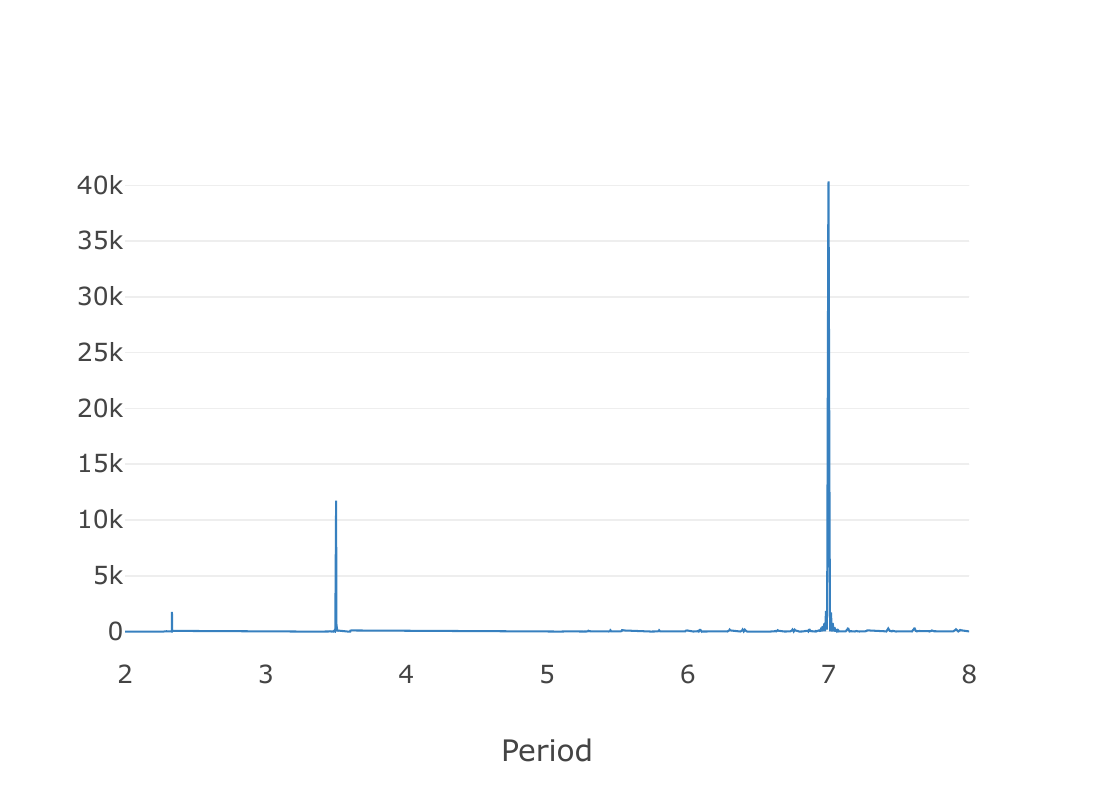}}
	\subfloat{\includegraphics[width=.48\textwidth]{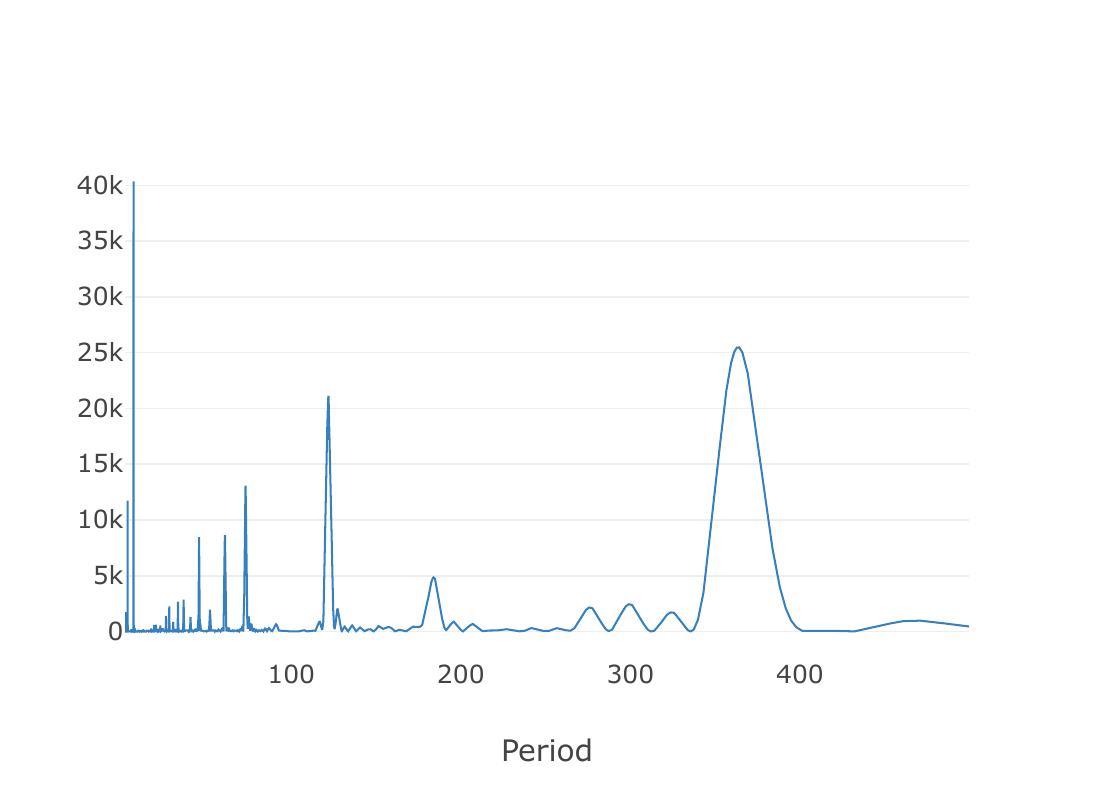}}
	\caption{IGD periodogram. Left panel: periods from 0 to 8 days; right panel: periods from 0 to 500 days.}
	\label{igd_periodogram}
\end{figure}

The periodogram, plotted in \cref{igd_periodogram}, exhibits peaks at periods of 365.25 and 7 days, while other relevant values are ascribable to multiple harmonics of the fundamental ones. Notably, differently from what happens for RGD \cite{fabbiani2019forecasting}, the weekly seasonality prevails on the yearly one in terms of magnitude. 

\begin{figure}[H]
	\centering
	\subfloat{\includegraphics[width=.48\textwidth]{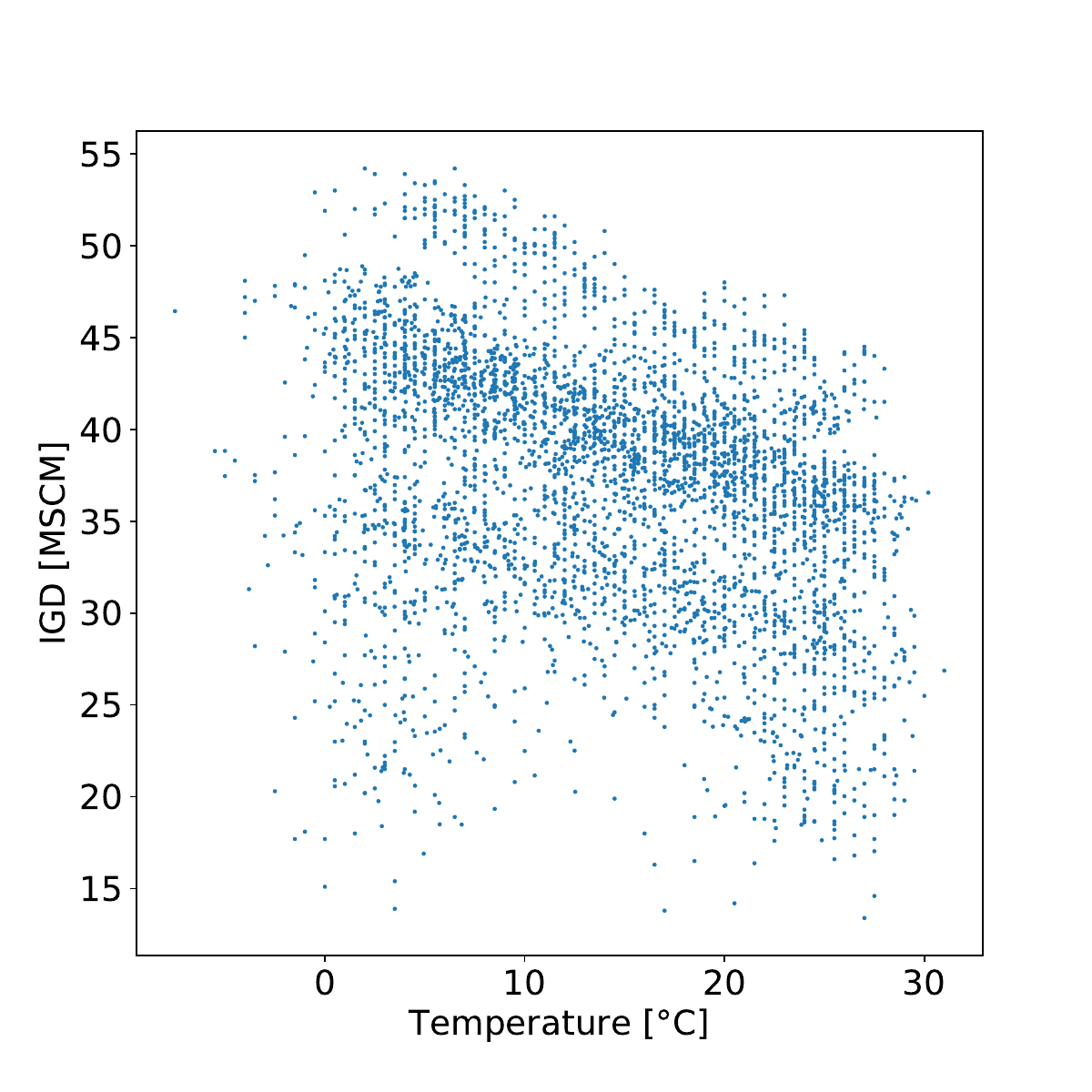}}
	\subfloat{\includegraphics[width=.48\textwidth]{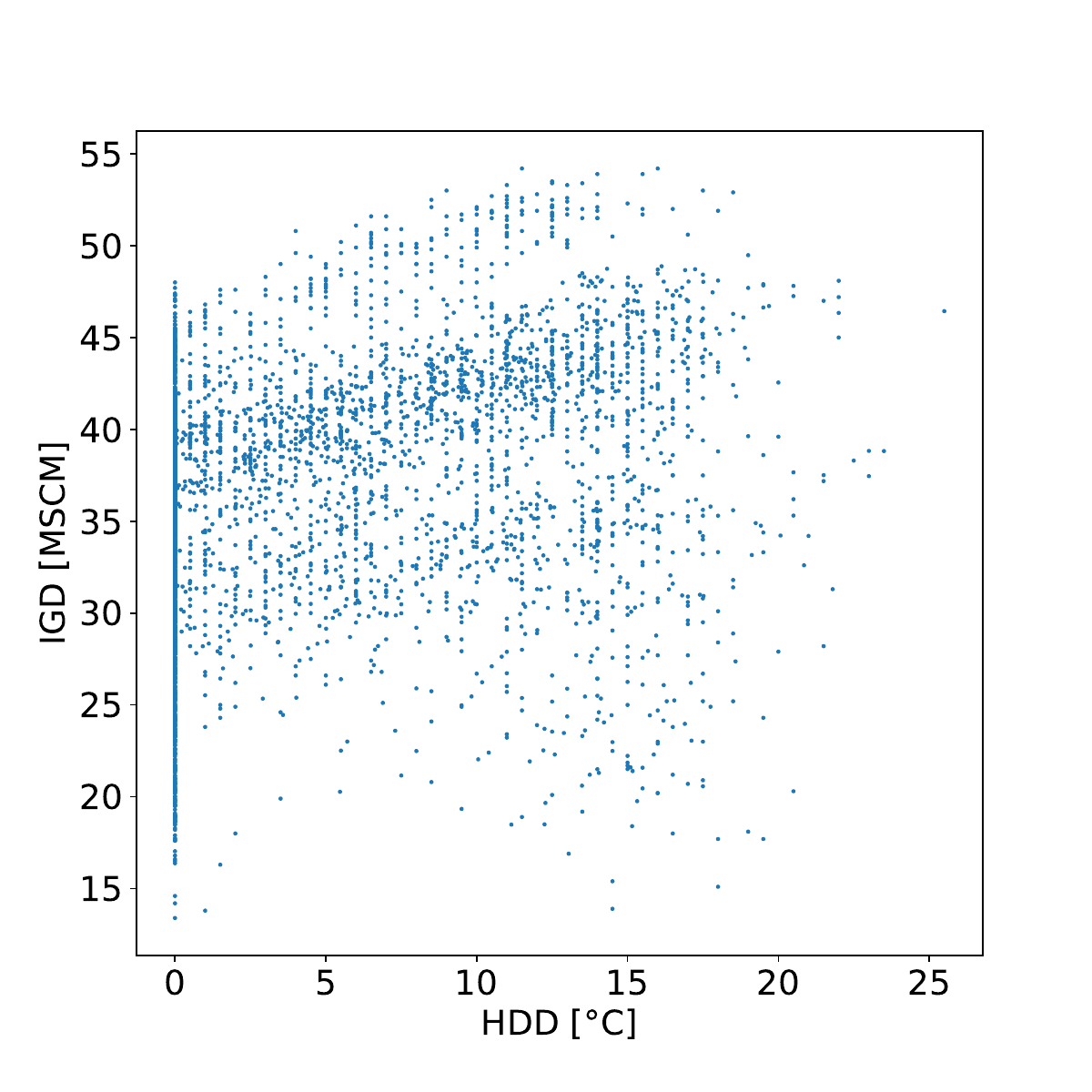}}
	\caption{Effect of temperature on IGD. Left panel: IGD vs temperature; right panel: IGD vs Heating Degree Days (HDD).}
	\label{igd_temp}
\end{figure}

Temperature is known to be a major determinant of gas demand \cite{vsebalj2017predicting, BALDACCI2016190, gil2004generalized}. In order to take into account that the need for heating ceases when temperature raises above $18\,^{\circ}{\rm C}$, it is useful to refer to the so-called Heating Degree Days (HDD), defined as $\mathrm{HDD} = \max(18-T, 0)$, where  $T$ is the temperature in degrees Celsius. The scatter plots of IGD against temperature and IGD against HDD are reported in \cref{igd_temp}.

\subsection{Thermoelectric gas demand}
Differently from IGD, Thermoelectric Gas Demand (TGD) shows a clear trend, see \cref{GD_vs_time}. From 2008 to 2014 TGD decreases, mostly due to the growing importance of renewable sources of electric power, while, since 2014, the trend stabilises, likely due to the decrease in subsidies to the installation of photovoltaic systems.

\begin{figure}[H]
    \centering			\includegraphics[width=\textwidth]{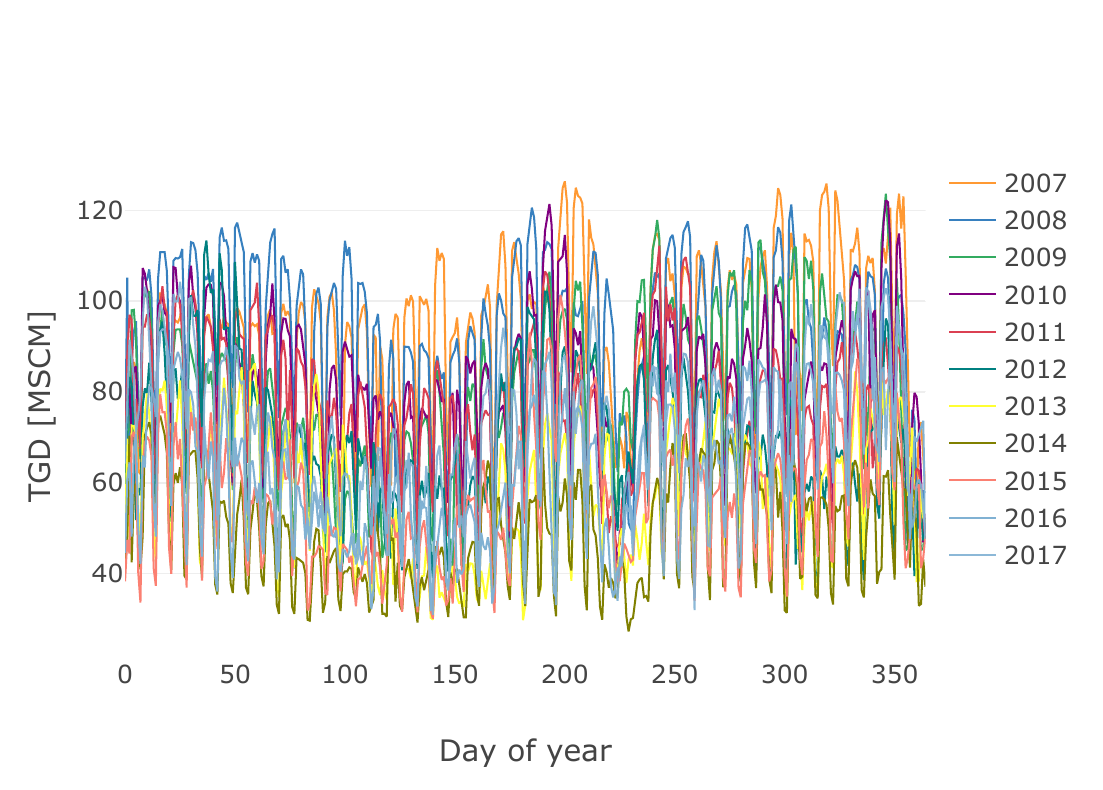}
    \caption{TGD, years 2007-2017. The time series has been shifted to align weekdays. \label{tgd_superimp}}
\end{figure}

\begin{figure}[H]
	\centering
	\subfloat{\includegraphics[width=.48\textwidth]{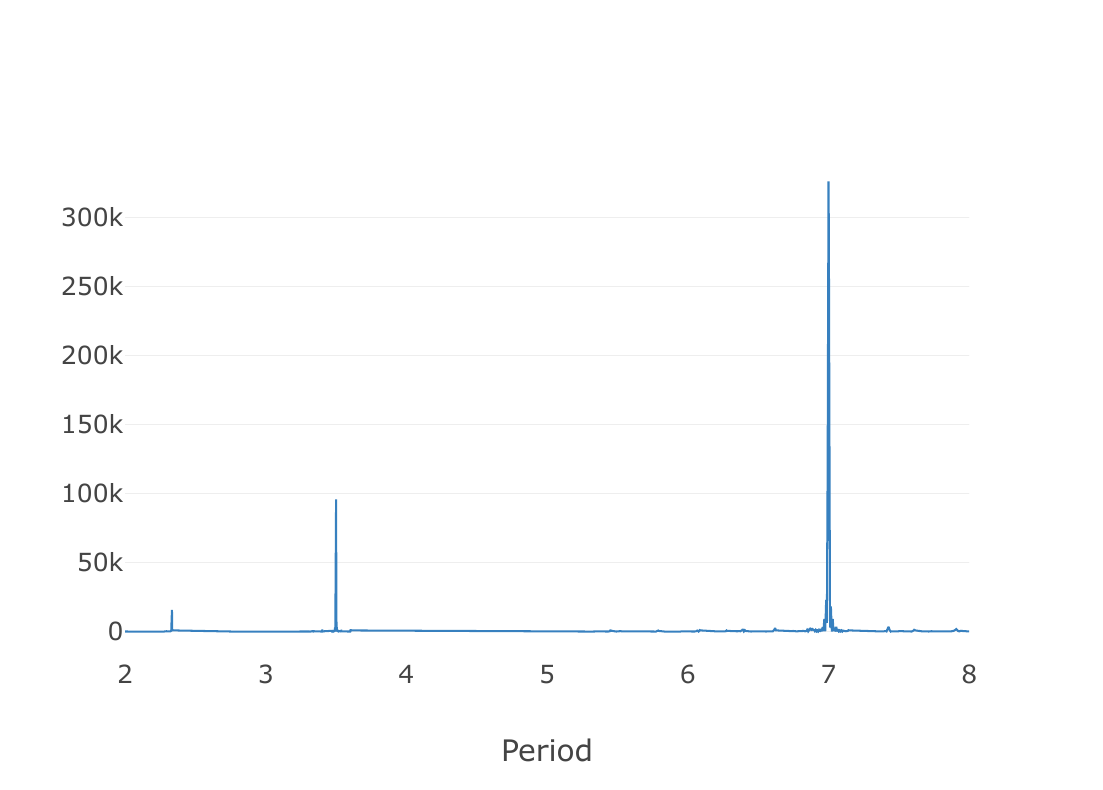}}
	\subfloat{\includegraphics[width=.48\textwidth]{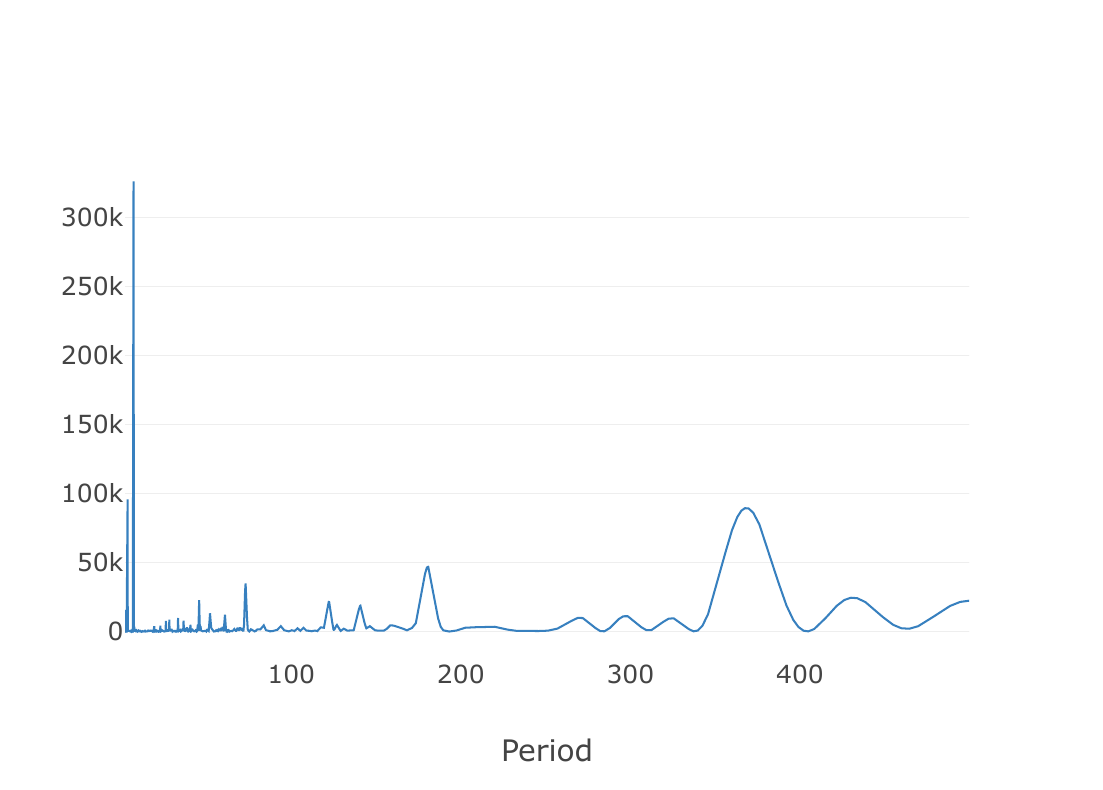}}
	\caption{TGD periodogram. Left panel: periods from 0 to 8 days; right panel: periods from 0 to 500 days}
	\label{tgd_periodogram}
\end{figure}

Moreover, TGD shows a greater variability compared to IGD and RGD, as seen from the year-over-year plot in \cref{tgd_superimp}. TGD is indeed influenced by several  factors, including prices of electric power, gas, and European Emission Allowance (EUA) certificates, which exhibit a large volatility \cite{weron2006modeling}. This  explains why yearly periodicity is relatively less important in TGD than in IGD and RGD.
The periodogram in \cref{tgd_periodogram} shows that, also for TGD, the main seasonal component is the weekly one, which is consistent with Italian power demand \cite{guerini2015long}.

\begin{figure}[H]
	\centering
	\subfloat{\includegraphics[width=.48\textwidth]{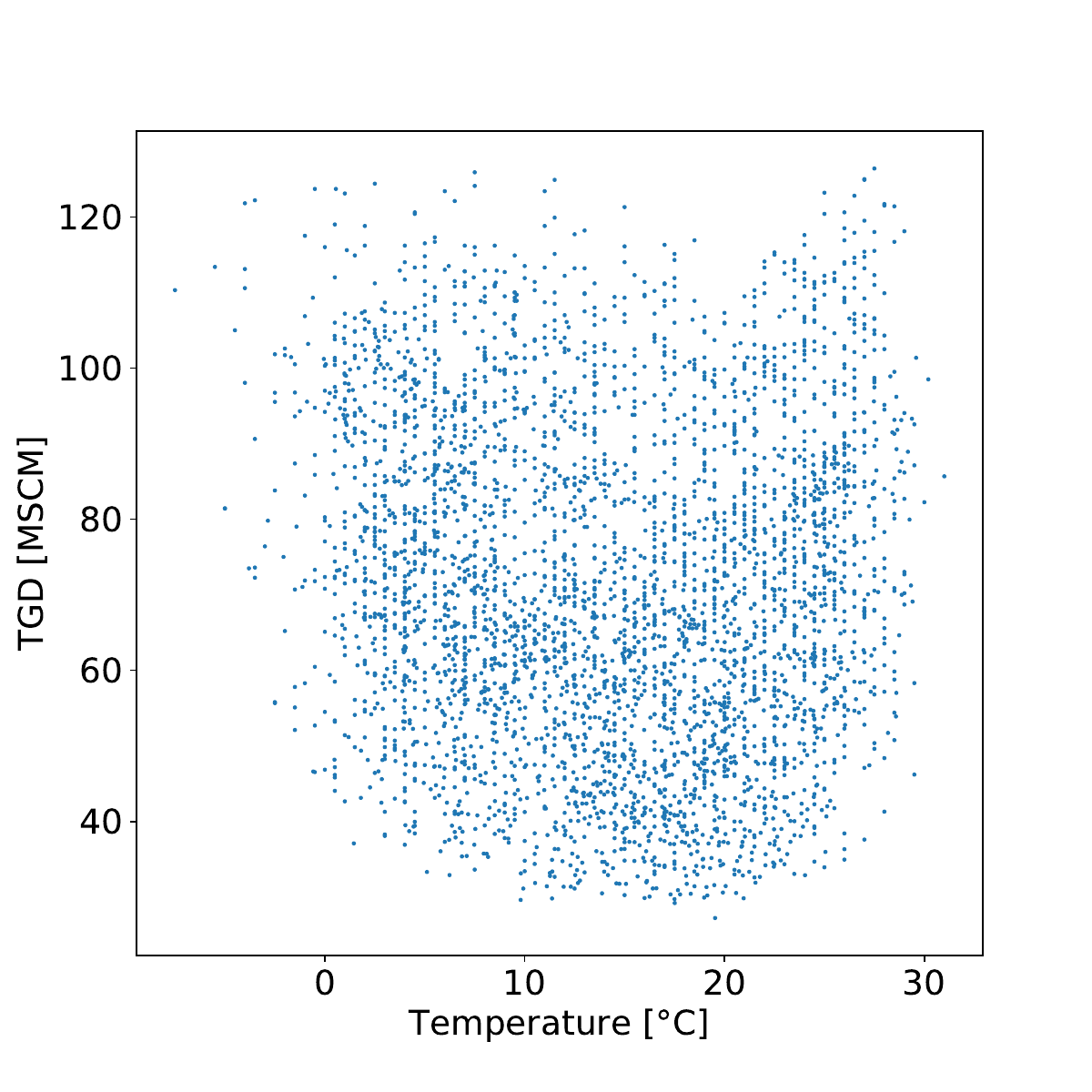}}
	\subfloat{\includegraphics[width=.48\textwidth]{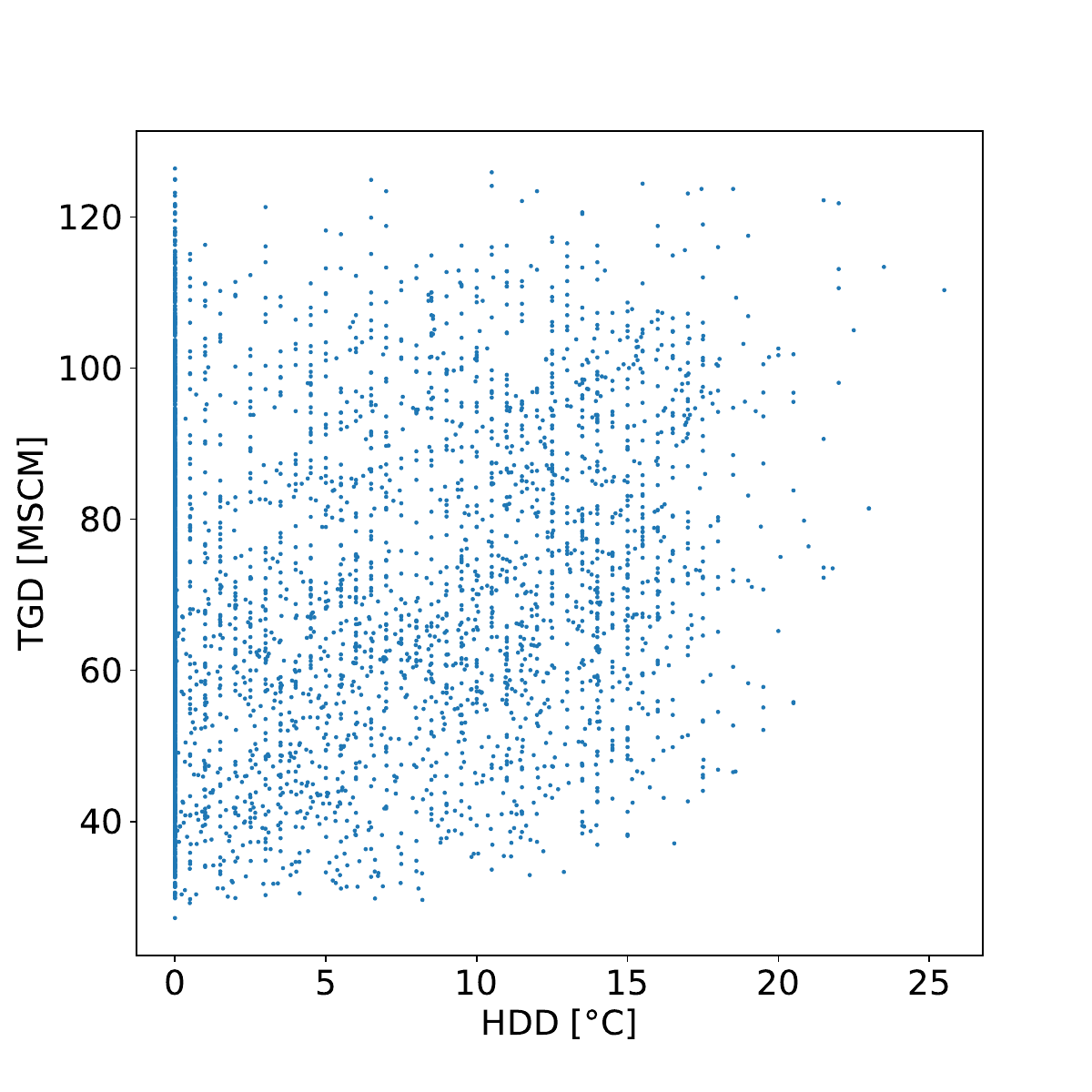}}
	\caption{Effect of temperature on TGD. Left panel: TGD vs temperature; right panel: TGD vs HCDD.}
	\label{tgd_temp}
\end{figure}

The scatter plot of TGD against temperature, displayed in the left panel of \cref{tgd_temp}, shows a peculiar U-shaped pattern: TGD increases as weather gets colder, but also when it gets hotter. In fact, in summer more thermoelectric production is required because air conditioning pushes the demand for electric power. This U-shaped pattern justifies the introduction of a suitable feature variable, herein named \emph{Heating and Cooling Day Degrees} (HCDD). More precisely
\begin{equation*}
    \mathrm{HCDD} = \left| T_c - T \right|
\end{equation*}
We found that $T_c=16°C$ maximises the linear correlation between TGD and HCDD.

\section{Feature extraction} \label{feature_selection}
Based also on the exploratory analysis, the features used in the prediction algorithms include: autoregressive terms, calendar features, temperature and its derived variables HDD and HCDD. \Cref{features} reports the complete list of features.

\subsection{Autoregressive features}
In order to capture the yearly periodicity and the concurrent effect of holidays, it is convenient to resort to the notion of similar day, see e.g. \cite{fabbiani2019forecasting}, here recalled for convenience. 
Letting $t$ and $\tau$ be calendar dates, we define:
\begin{itemize}
\item $\mathrm{year}(t)$: the year to which day $t$ belongs;
\item $\mathrm{weekday}(t)$: the weekday of day $t$, e.g. Monday, Tuesday, etc;
\item $\mathrm{yearday}(t)$: the day number within $\mathrm{year}(t)$ starting from January 1, whose \emph{yearday} is equal to $1$.
\end{itemize}
\begin{definition}[Similar Day \cite{fabbiani2019forecasting}] 
If $t$ is not a holiday\footnote{According to the Italian calendar, holidays are: 1 January, 6 January, 25 April, 1 May, 2 June, 15 August, 1 November, 8, 25 and 26 December, Easter and Easter Monday.}, its similar day $\tau^*=\mathrm{sim}(t)$ is
$$
\tau^* = \arg \min_{\tau} |\mathrm{yearday}(\tau)-\mathrm{yearday}(t)|
$$
subject to
\begin{itemize}
\item $\mathrm{year}(\tau)=\mathrm{year}(t)-1$;
\item $\mathrm{weekday}(\tau)=\mathrm{weekday}(t)$;
\item $\tau$ is not a holiday.
\end{itemize}
If $t$ is a holiday, its similar day $\tau^*=\mathrm{sim}(t)$ is the same holiday in the previous year.
\end{definition}

To predict $y_t, y \in \{\mathrm{IGD}, \mathrm{TGD}\}$, we included as features $y_{t-1}$, $y_{t-7}$, $y_{\mathrm{sim}(t)}$ and $y_{\mathrm{sim}(t-1)}$.

\subsection{Calendar features}
As calendar features, we introduced binary dummy variables to account for weekdays and holidays. Dummy variables were also added to identify: (i) extended holidays, i.e. working days preceded and followed by either Saturdays, Sundays or holidays, and (ii) days after holidays, i.e. working days which immediately follow a holiday and are not extended holidays.

\subsection{Temperature features}
We selected as features also forecasted temperatures $T_t$, $T_{t-1}$, $T_{t-7}$ and $T_{\mathrm{sim}(t)}$. In view of what shown in \cref{exploratory_analisys}, for IGD, also HDD values at the same times were introduced, while, for TGD, HCDD replaced HDD.

\begin{table}[]
\centering
\begin{tabular}{@{}llll@{}}
\toprule
Feature                & Reference time & Type        & Series\\ \midrule \addlinespace
Gas demand series                    & t-1            & continuous   & RGD, IGD, TGD\\
Gas demand series                    & t-7            & continuous   & RGD, IGD, TGD\\
Gas demand series                    & $\mathrm{sim}(t)$         & continuous  & RGD, IGD, TGD\\
Gas demand series                    & $\mathrm{sim}(t-1)$       & continuous  & RGD, IGD, TGD\\
\addlinespace
Forecasted temperature & t              & continuous  & RGD, IGD, TGD\\
Forecasted temperature & t-1            & continuous  & RGD, IGD, TGD\\
Forecasted temperature & t-7            & continuous  & RGD, IGD, TGD\\
Forecasted temperature & $\mathrm{sim}(t)$         & continuous  & RGD, IGD, TGD\\
Forecasted HDD         & t              & continuous  & RGD, IGD\\
Forecasted HDD         & t-1            & continuous  & RGD, IGD\\
Forecasted HDD         & t-7            & continuous  & RGD, IGD\\
Forecasted HDD         & $\mathrm{sim}(t)$         & continuous  & RGD, IGD\\ 
Forecasted HCDD         & t              & continuous  & TGD\\
Forecasted HCDD         & t-1            & continuous  & TGD\\
Forecasted HCDD         & t-7            & continuous  & TGD\\
Forecasted HCDD         & $\mathrm{sim}(t)$         & continuous  & TGD\\ 
\addlinespace
Weekday                & t              & categorical       & RGD, IGD, TGD\\
Holiday                & t              & dummy    & RGD, IGD, TGD\\
Day after holiday      & t              & dummy    & RGD, IGD, TGD\\
Bridge holiday         & t              & dummy    & RGD, IGD, TGD\\   
\bottomrule
\end{tabular}
\caption{Features \label{features}}
\end{table}

\section{Predictive models} \label{predictive_models}

We selected nine base models, which can be grouped into three categories: 
\begin{enumerate}
    \item linear models: ridge regression, lasso, Torus model \cite{guerini2015long}, support vector regression, and elastic net
    \item non-linear models: random forest, neural networks
    \item non-parametric models: Gaussian Process, nearest neighbour
\end{enumerate}

Four of them, namely ridge regression, Gaussian Process (GP), Torus model, nearest neighbours and neural networks, were already applied to RGD: we thus refer to \cite{fabbiani2019forecasting} for their description.

Moreover, we introduced four ensemble models, described in \cref{2pm}, which aggregate forecasts issued by a subset of the basic models: (i) simple average, (ii) average on an optimised subset of base forecasts (subset average), (iii) weighted average, (iv) support vector regression.

\subsection{Base models}
For the design of the predictors, we assume that $n$ training data pairs $(\textbf{x}_i,y_i)$, $i=1, \ldots, n$, are used to devise a prediction rule $f(\cdot)$. In particular, letting $(\textbf{x}_*,y_*)$ denote a novel input-output pair, $f(\textbf{x}_*)$ will be used to predict $y_*$. Herein, $\textbf{x}_i \in \mathbb{R}^p$, $p<n$, is a vector whose entries are given by the $p$ features associated with the target $y_i$. 

In the following, $\textbf{y}=y_i \in \mathbb{R}^n$ is the vector of the training target data, while $\textbf{X}= \{x_{ij}\} \in \mathbb{R}^{n \times p}$ is the matrix of the training input data, where $x_{ij}$ is the $j$-th feature of the $i$-th training pair $(\textbf{x}_i,y_i)$.

\subsubsection*{Linear models}

Ridge regression \cite{hoerl1970ridge}, LASSO \cite{tibshirani1996regression} and elastic net \cite{zou2005regularization} are  methods to identify the parameters $\beta_j$ of the linear-in-parameter predictor:
\begin{equation*}
f(\textbf{x}) = \sum_{j=1}^{p}x_{j}\beta_j = \textbf{x}^{T}  \boldsymbol{\beta},  \quad \boldsymbol{\beta} \in \mathbb{R}^p
\end{equation*}
where $\mathbf{x}^T=\left[\begin{array}{cccc}x_1 & x_2 & \ldots & x_n\end{array}\right]$. Accordingly, the vector of the predicted training targets is
\begin{equation}
\mathbf{f}= \left[\begin{array}{cccc}f(\mathbf{x_1}) & f(\mathbf{x_2}) & \ldots & f(\mathbf{x_n})\end{array}\right]^T = \mathbf{X}\boldsymbol{\beta}
\label{linear_model}
\end{equation}

To prevent overfitting and improve generalization capabilities, in all the three methods the loss function includes a penalty on the magnitude of $\boldsymbol \beta$:
\begin{align*}
	\boldsymbol{\beta}^{\text{ridge}} &:= \arg \min_{\boldsymbol{\beta}} \| \mathbf{y-X} \boldsymbol{\beta} \|^2 +\lambda \|\boldsymbol{\beta} \|^2 \\
	\boldsymbol{\beta}^{\text{LASSO}} &:= \arg \min_{\boldsymbol{\beta}} \| \mathbf{y-X} \boldsymbol{\beta} \|^2 +\lambda \sum_{i=1}^p |\beta_i | \\
	\boldsymbol{\beta}^{\text{elastic net}} &:= \arg \min_{\boldsymbol{\beta}} \| \mathbf{y-X} \boldsymbol{\beta} \|^2 +\lambda \left( \alpha \| \boldsymbol{\beta} \|^2 + (1 - \alpha) \sum_{i=1}^p |\beta_i | \right)
\end{align*}

The three methods share the same standard quadratic loss
$$
\| \mathbf{y-X} \boldsymbol{\beta} \|^2 = \sum_{i=1}^n L(y_i,f(\mathbf{x_i})), \quad L(y, \hat{y}) := (y - \hat{y})^2
$$
but use different penalties that result in specific shrinking patterns. In ridge regression, the quadratic penalty shrinks parameters toward the origin. In the LASSO the penalty on the sum of absolute values has the effect of zeroing the least relevant parameters, thus enforcing some degree of sparsity. In the elastic net an intermediate effect is achieved.

Both $\lambda$ and $\alpha$ play the role of hyperparameters: $\lambda$ controls the strength of the parameter shrinkage, while $\alpha$ tunes the balance between penalties on the $L_1$ and the $L_2$ norm of $\boldsymbol \beta$. 

An alternative method to estimate $f$ is by means of support vector regression (SVR), see e.g. \cite{smola2004tutorial}. In this case, in place of the quadratic loss, an $\epsilon$-insensitive loss function is used:
\begin{equation}
    L_{\epsilon}(y,\hat{y}):=
    \begin{cases}
    0, & \abs{y-\hat{y}} < \epsilon  \\
    \abs{y-\hat{y}} -\epsilon, & \text{otherwise}
   \end{cases}
\end{equation}
Moreover, the assumption is made that  $f \in \mathcal{H}$, where $\mathcal{H}$ is a Reproducing Kernel Hilbert Space (RKHS) \cite{evgeniou2000regularization}, whose reproducing kernel is denoted by $K(\cdot, \cdot)$. Under this assumption,
$$
\| f \|_K = \sum_{j=1}^\infty \sum_{i=1}^\infty \alpha_j \alpha_i K(x_j,x_i)
$$
where ${\alpha_i}$ are such that
$$
f(\mathbf x )=\sum_{j=1}^\infty \alpha_j K(\mathbf x,\mathbf{x_j})
$$
The SVR estimate is defined as

\begin{equation}
 f^{\text{SVR}} := \arg \min_{f \in \mathcal{H}} \sum_{i=1}^{n}L_{\epsilon}(y_i, f(\mathbf{x_i})) + \frac{\lambda}{2} \| f\|_K^2  
\end{equation}
The hyperparameters are the real-valued constants $\lambda$ and $\epsilon$.

Although $f^{\text{SVR}}(\mathbf{x})$ is a nonlinear function of $\mathbf{x}$, the \emph{Representer Theorem} (see e.g. \cite{smola2001representer}) ensures that there exist coefficients $c_i$ such that the predictor function can be written as a linear combination of kernel functions centered at $\mathbf{x_i}$
$$
f^{\text{SVR}}(\mathbf{x}) = \sum_{i=1}^n c_i K(\mathbf{x_i},\mathbf{x})
$$
so that also the SVR predictor, though implementing a nonlinear function of the features, has a linear-in-parameter structure.

\subsubsection*{Nonlinear models: Random forest}

The Random forest method (see e.g. \cite{murphy2012machine}) is based on so-called Classification and Regression Trees (CART) \cite{breiman2017classification}. CARTs perform a recursive feature-wise partitioning of the input space and fit local linear regressions in each region of the final partition. CARTs are known to be unstable and prone to overfitting. In order to overcome these limitations, random forest models grow multiple CARTs, resorting to so-called data and feature bagging.  
Bagging or bootstrap aggregating is a random selection of a subset of a dataset that is repeated multiple times (with replacement). Models are then trained on each selected subset. By applying bagging to both data and features, each tree gets trained on different samples and feature sets. Forecasts performed by all models are then averaged to get the final prediction, leading to a more stable model.

\subsection{Ensemble models} \label{2pm}

Four alternative aggregation techniques were considered.
Apart from the Simple average, the calibration of the other three methods requires a specific \emph{ensemble training dataset} not used for training the base models.

\emph{Simple average}. The most trivial aggregation is the arithmetic average of the forecasts achieved by base models. Given a test input $\mathbf{x}_*$, and $M$ base forecasts $\hat{f}_i(\mathbf{x}_*)$, $i=1, ..., M$, the ensemble forecast is

\begin{equation*}
    \hat f^{\mathrm{A}}(\mathbf{x_*})=\frac{1}{M}\sum_{i=1}^{M}\hat f_{i}(\mathbf{x_*})
\end{equation*}

\emph{Weighted average}. A second option is the weighted average of base forecasts:

\begin{equation*}
    \begin{cases}
        \hat f^{\mathrm{LS}}(\mathbf{x_*})=\sum_{i=1}^{M}w_{i}\hat f_{i}(\mathbf{x_*})  \\
        w_{i}\geq0 \\
        \sum_{i=1}^{M}w_{i}=1
   \end{cases}
\end{equation*}

The weights $w_{i}$ are obtained by minimizing the sum of squared residuals between the ensemble forecast and the target vector on the ensemble training dataset.

\emph{Subset average}. The third ensemble method computes the average of a suitable subset of predictors. The chosen subset is obtained by a brute force search within the set of all possible subsets, choosing the subset of predictors whose average minimises the MAE computed on the ensemble training dataset. In our case, excluding the complete subset made of all the nine predictors (already considered as simple average), and the nine base models, the number of candidate subsets is
\begin{equation*}
    \sum_{k=2}^8 {\binom{9}{k}} = 501 
\end{equation*}

\emph{SVR aggregation}. The fourth ensemble method trains a SVR model on the ensemble training dataset, using base forecasts as features.

\section{Implementation notes} \label{experiments}
The available data range from 2007 to 2018. Four one-year long test sets, ranging from 2015 to 2018, were used to obtain a comparative assessment of the 13 models, including 9 base models and 4 ensemble ones. Each test set was associated to a set of training data, that were organised differently depending on the nature of the considered model, either base or ensemble.

\emph{Training of base models}. The training set, called \emph{base training set} $\mathcal T_{\mathrm{base}}(\mathcal Y)$, is made of all data previous to the test year $\mathcal Y$. For instance, if $\mathcal Y=\{2017\}$ is taken as test set, the 9 base models were trained on the \emph{base training set} $\mathcal T_{\mathrm{base}}(\{2017\})=\{2007, \ldots, 2016\}$.

\emph{Training of ensemble models}. In this case two training sets were considered. The year before the test set $\mathcal Y$ was used as \emph{ensemble training set} $\mathcal T_\mathrm{ens}(\mathcal Y)$, while the remaining data were used to train the 9 base models that enter the aggregation. For instance, if $\mathcal Y=\{2017\}$ is taken as test set, the 9 base models were trained on $\mathcal T_\mathrm{base}(\{2016\})=\{2007, \ldots, 2015\}$, while the ensemble models were trained on $\mathcal T_\mathrm{ens}(\{2017\})=\{2016\}$.

Hyperparameters of the Torus model were tuned by maximising AIC, those of the Gaussian Process by maximising the marginal likelihood, while for all the other base models five-fold cross validation was used.

\emph{Test of base models}. For the test set $\mathcal Y$, base predictions were computed using the base models trained on $\mathcal T_{\mathrm{base}}(\mathcal Y)$. 

\emph{Test of ensemble models}. Given the forecasts provided by the base models trained on $\mathcal T_{\mathrm{base}}(\mathcal Y)$, the ensemble forecasts were obtained using ensemble models trained on $\mathcal T_\mathrm{ens}(\mathcal Y)$.

Out-of-sample performances were evaluated in terms of Mean Absolute Error (MAE): 

\begin{equation*}
    \mathrm{MAE} = \frac{1}{n}\sum\limits_{t=1}^n |y_t-\hat{y}_t|
\end{equation*}

where $y_t$ and $\hat{y}_t$, $y \in \{\mathrm{RGD}, \mathrm{IGD}, \mathrm{TGD}, \mathrm{GD} \}$, are the actual value and its forecast, while $n$ is the number of samples in the considered test set. MAE was preferred over percent or relative error metrics due to the large range of values assumed by the target variable, which would give undue importance to poor performances during low-demand periods.   

\section{Results} \label{results}

Mean absolute errors for RGD, IGD, TGD, and total GD are reported, respectively, in \cref{rgd}, \cref{igd}, \cref{tgd} and \cref{gd}.

For what concerns base models, GP, ANN and SVR achieved the best average MAE across all the gas demands, with differences between each other smaller than 0.10 MSCM. Notably, results achieved by such models were also stable across different test sets. On the other hand, KNN was consistently the worst performer, due to its poor capability of modelling influence of temperature and holidays.

Ensemble models consistently outperformed base ones. In particular, subset average achieved the best average MAE on all four types of gas demand: the three disaggregated demands and the total one. A possible explanation is that different models are better at capturing specific behaviours: in \cite{fabbiani2019forecasting}, for instance, it was shown that the ANN model achieved the best results in winter, while GP in summer, suggesting that the former is better at modelling the impact of weather, while the latter can better follow seasonal patterns. Aggregation can indeed mitigate errors committed by single models, thus increasing overall accuracy and robustness. 

\begin{figure}[H]
    \centering
	\subfloat{
	    \includegraphics[width=.90\textwidth]{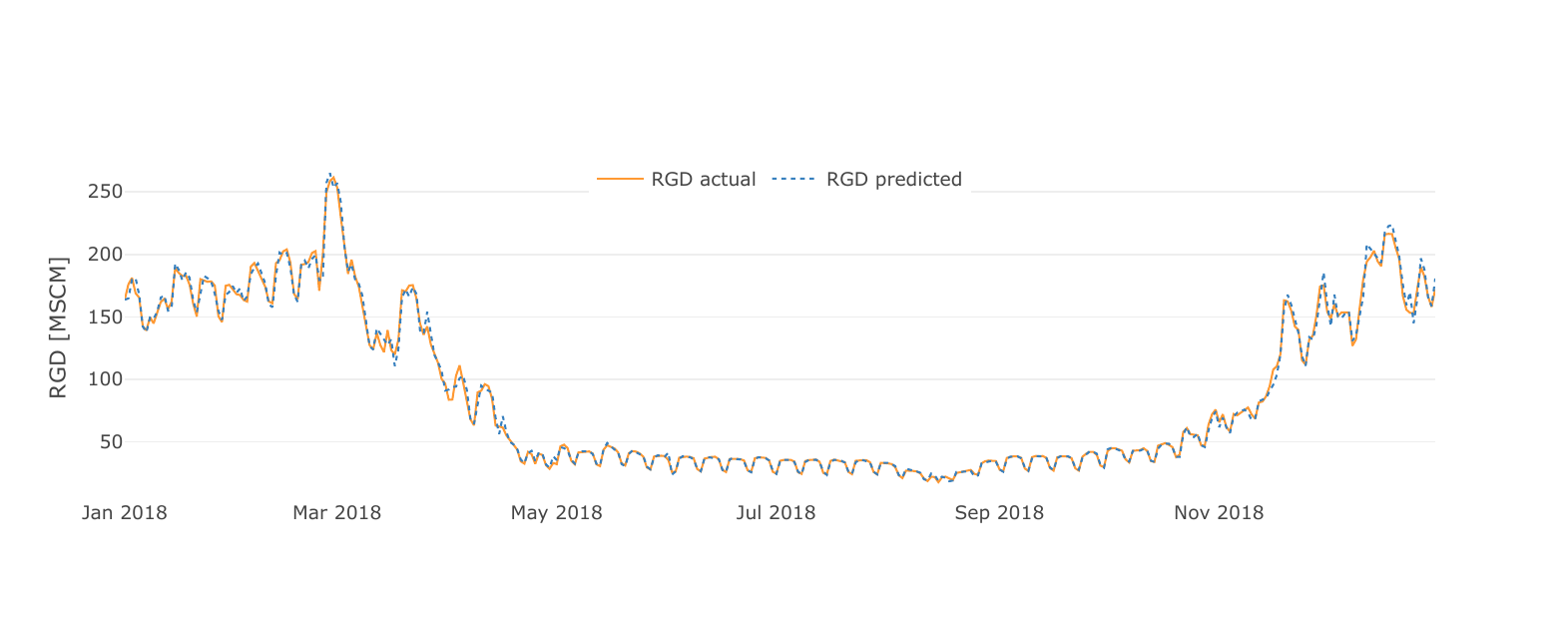}
	 } \\
	\subfloat{
	    \includegraphics[width=.90\textwidth]{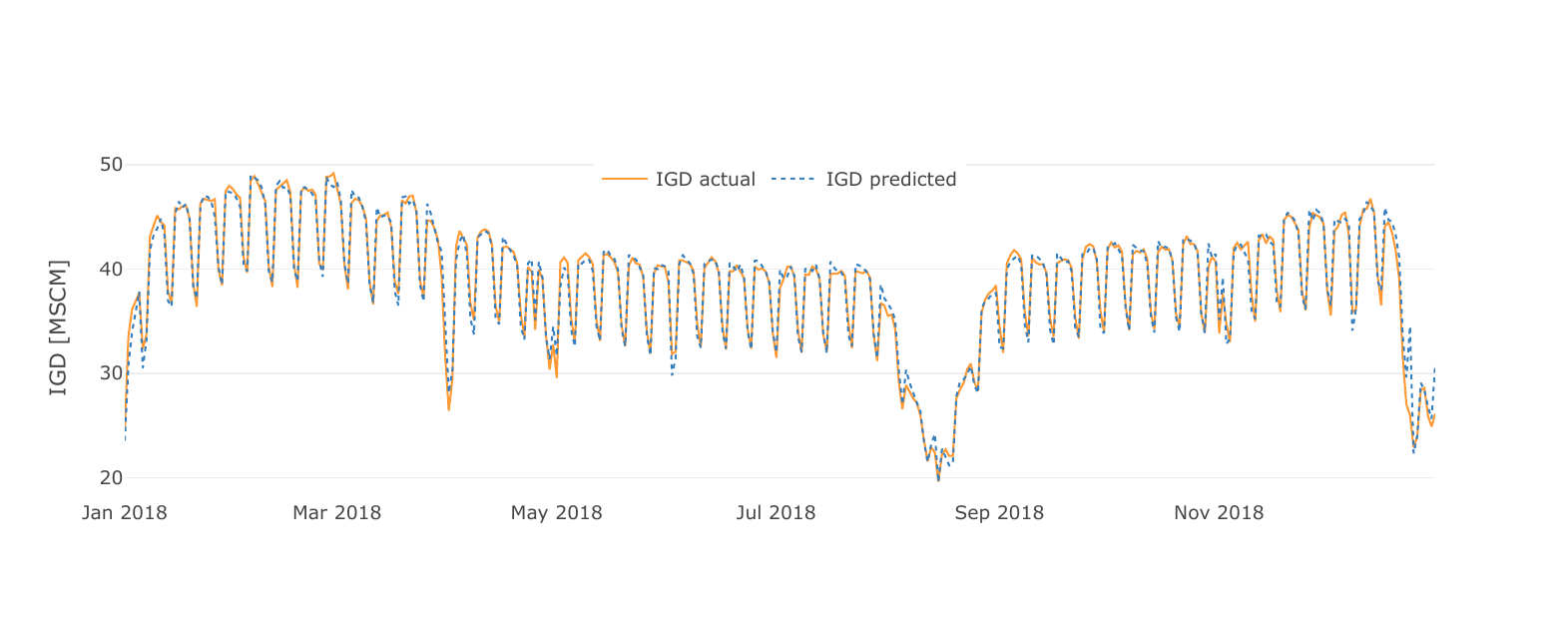}
	 } \\
	 \subfloat{
	    \includegraphics[width=.90\textwidth]{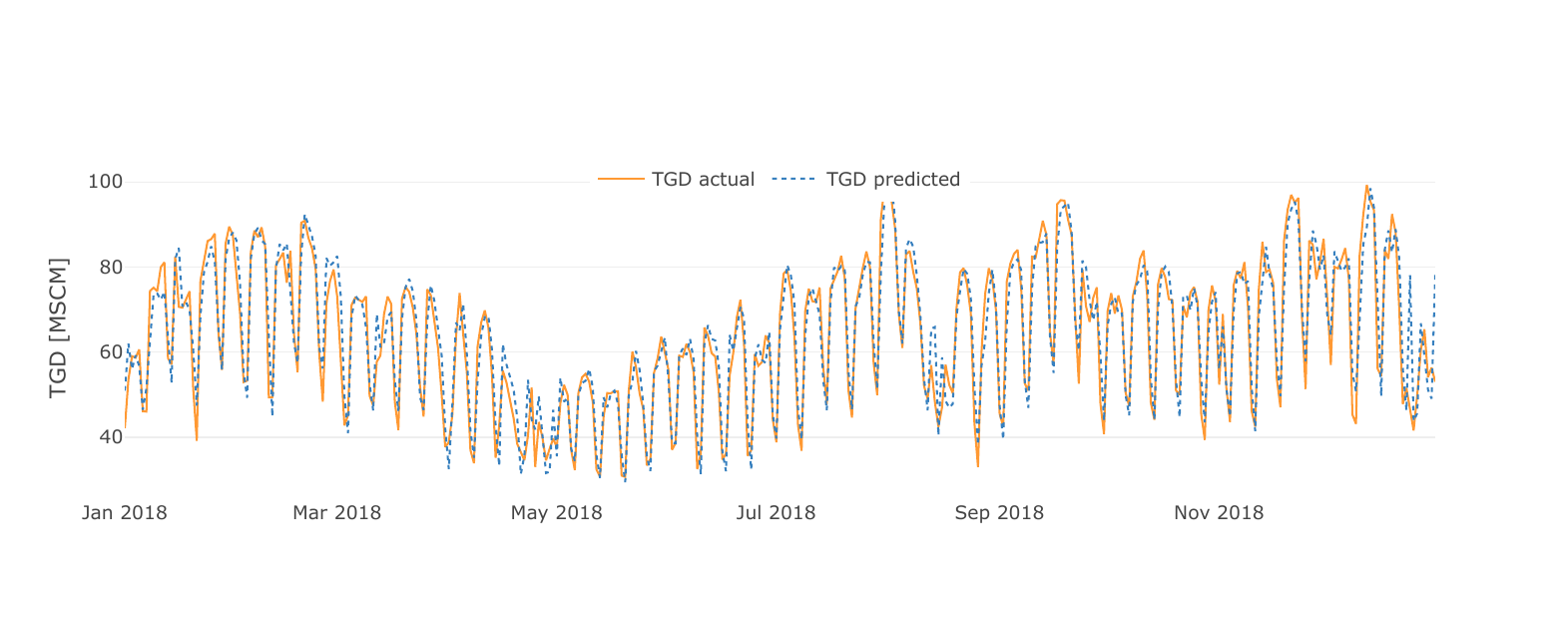}
	 } \\
	\subfloat{
	    \includegraphics[width=.90\textwidth]{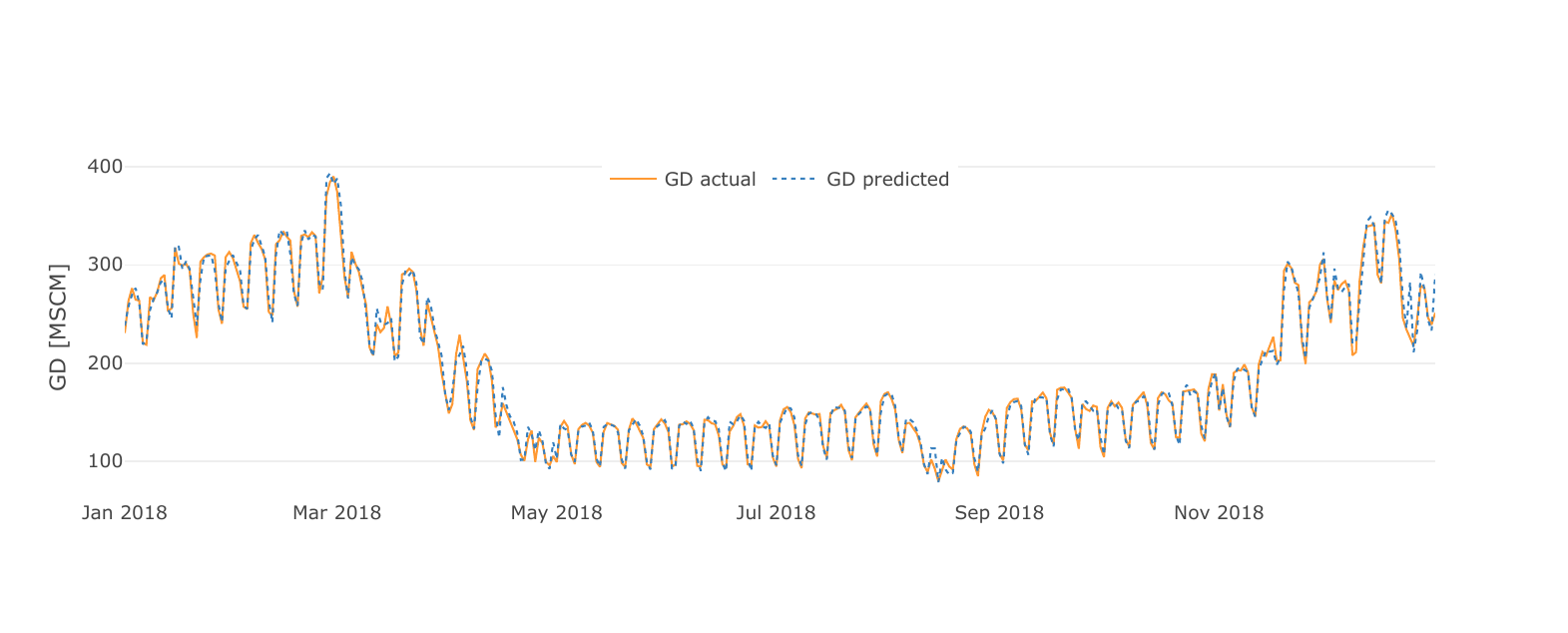}
	 }
	\caption{Subset average ensemble model: predicted gas demands in 2018. From top to bottom:  Residential Gas Demand (RGD), Industrial Gas Demand (IGD), Thermoelectic Gas Demand (TGD), overall Gas Demand (GD).}
    \label{forecast_series}
\end{figure}

\begin{figure}[H]
    \centering
	\subfloat{
	    \includegraphics[width=.90\textwidth]{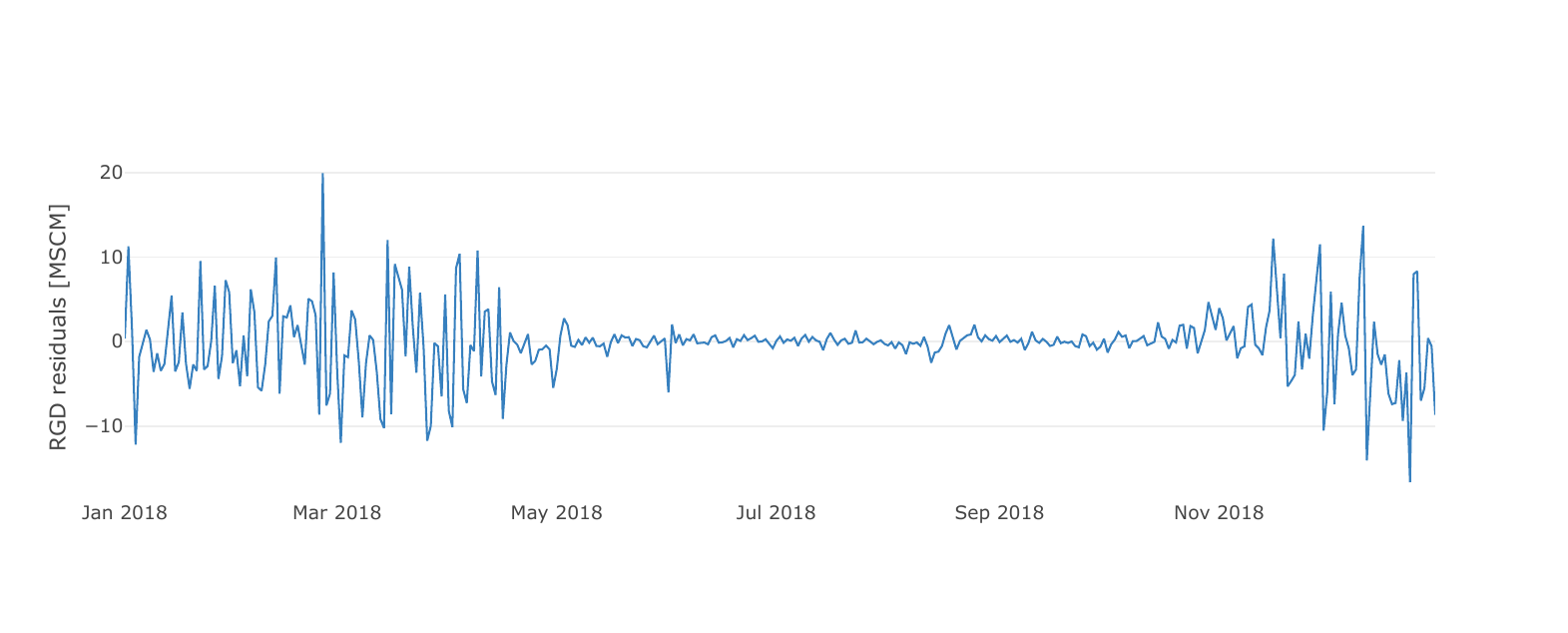}
	 } \\
	\subfloat{
	    \includegraphics[width=.90\textwidth]{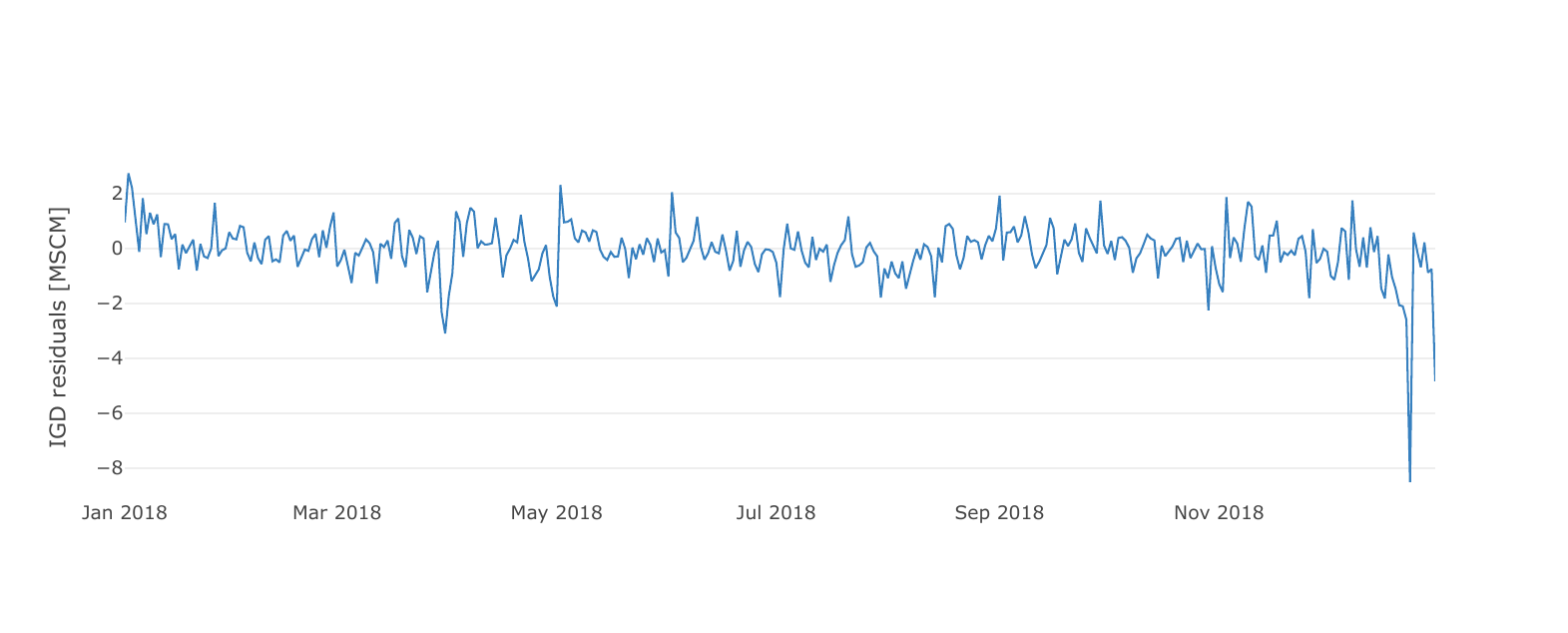}
	 } \\
	 \subfloat{
	    \includegraphics[width=.90\textwidth]{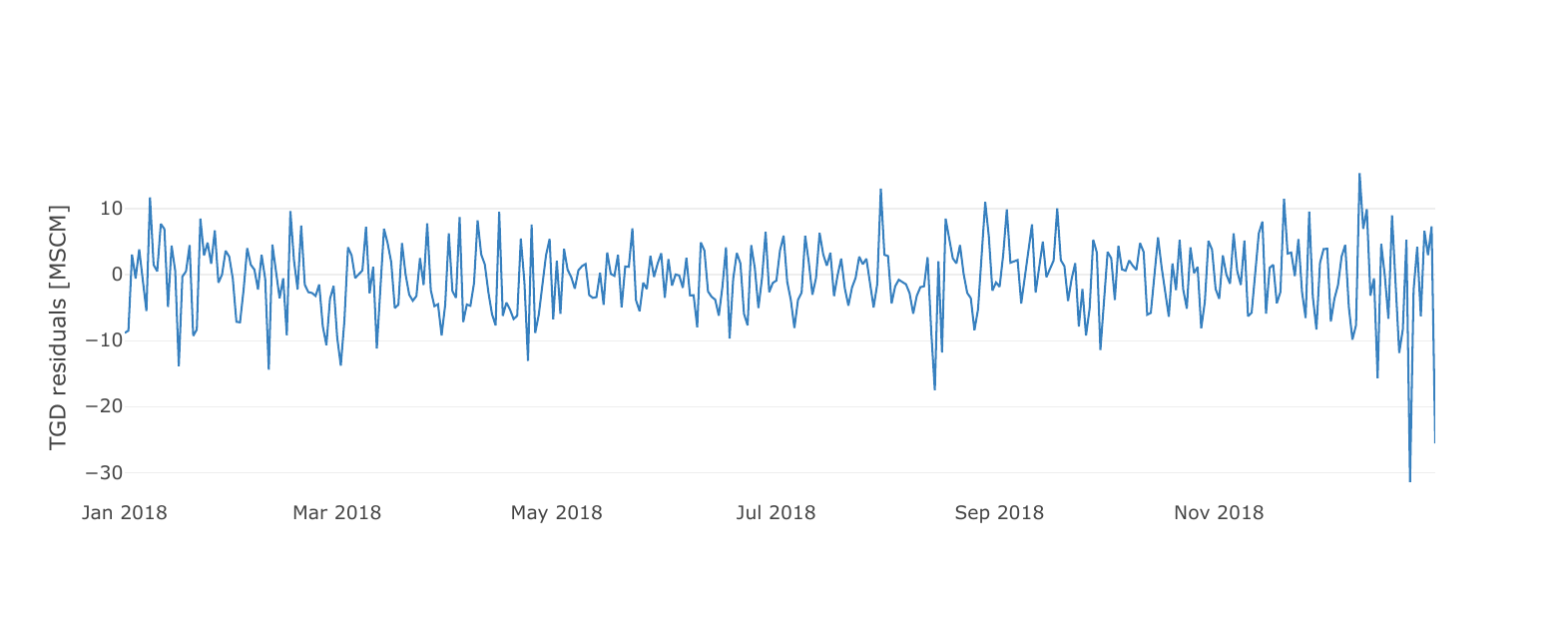}
	 } \\
	\subfloat{
	    \includegraphics[width=.90\textwidth]{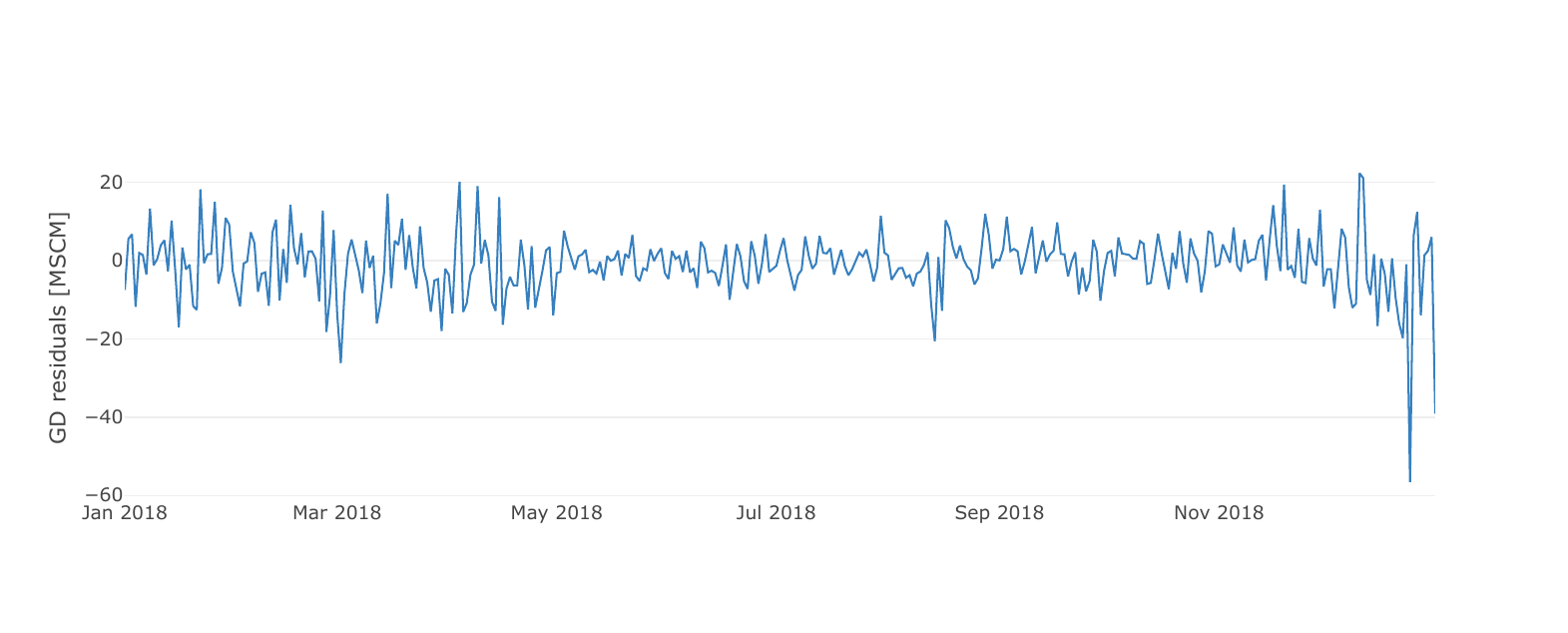}
	 }
	\caption{Subset average ensemble model: one-day-ahead prediction residuals in 2018. From top to bottom:  Residential Gas Demand (RGD), Industrial Gas Demand (IGD), Thermoelectic Gas Demand (TGD), overall Gas Demand (GD).}
    \label{forecast_residuals}
\end{figure}

The improvement due to aggregation was particularly evident for RGD (\cref{rgd}), where the best base model (GP) was outperformed by the best ensemble model (subset average) by 0.25 MSCM. The gap between base and ensemble models was smaller for the other two gas demands: GP and SVR are worse than subset average by 0.07 MSCM, for IGD (\cref{igd}); SVR is worse than subset average by 0.09 MSCM, for TGD (\cref{tgd}). Finally SVR is worse than subset average by 0.29 MSCM, for the global Italian GD (\cref{gd}). The 2018 forecasts and the corresponding residuals provided by the best ensemble predictor, namely subset average, are displayed in \cref{forecast_series} and \cref{forecast_residuals}, respectively.

To the best of our knowledge, the only term of comparison available for the task addressed in this paper is given by the forecasts of the global Italian GD issued by SNAM Rete Gas, the Italian Transmission System Operator (TSO) \cite{snamForecastData}. In 2017 and 2018, the improvement is neat: the out-of-sample MAE of SNAM predictions was 9.62 MSCM in 2017 and 8.30 MSCM in 2018, while our best model (subset average) scored 5.16 MSCM in 2017 and 5.46 MSCM in 2018, see \cref{gd}.

\begin{table}[H]
\centering
\begin{adjustbox}{max width=\textwidth}
\begin{tabular}{@{}lccccc@{}} 
\toprule
 Model  &  2015 &  2016 & 2017 & 2018 & Average  \\
\midrule
Ridge  & 3.39 & 3.10 & 3.01 & 3.49 & 3.25 \\
Lasso & 3.38 & 3.10 & 3.01 & 3.49 & 3.25 \\
Elastic net  & 3.38 & 3.10 & 3.01 & 3.49 & 3.25 \\
SVR  & 2.84 & 2.62 & 2.38 & 2.93 & 2.69 \\
GP & 2.60 & 2.48 & 2.51 & 2.61 & 2.55 \\
KNN & 4.57 & 5.51 & 5.08 & 5.52 & 5.17 \\
Random forest & 3.04 & 3.36 & 3.50 & 3.48 & 3.35 \\
Torus  & 3.18 & 2.66 & 2.54 & 3.13 & 2.88 \\
ANN & 2.76 & 2.68 & 2.43 & 3.10 & 2.74 \\
\addlinespace
Simple average  & 2.66 & 2.57 & 2.45 & 2.91 & 2.65 \\
Subset average  & \textbf{2.41} & \textbf{2.17} & \textbf{2.06} & \textbf{2.56} & \textbf{2.30} \\
Weighted average & 2.59 & 2.33 & \textbf{2.06} & 2.64 & 2.40 \\
SVR aggregation & 2.58 & 2.30 & 2.19 & 2.67 & 2.44 \\
\bottomrule
\end{tabular}
\end{adjustbox}
\caption{Forecasted Residential Gas Demand: out-of-sample MAEs. Each year's best performers are in boldface.}
\label{rgd}
\end{table}

\begin{table}[H]
\centering
\begin{adjustbox}{max width=\textwidth}
\begin{tabular}{@{}lcccccc@{}} 
\toprule
Model  &   2015 &   2016 &  2017 &  2018 & Average  \\
\midrule
Ridge  & 0.75 & 0.75 & 0.74 & 0.77 & 0.75 \\
Lasso & 0.75 & 0.75 & 0.74 & 0.77 & 0.75 \\
Elastic Net  & 0.75 & 0.75 & 0.74 & 0.77 & 0.75 \\
SVR  & 0.57 & 0.58 & 0.7 & 0.75 & 0.65 \\
GP & 0.61 & 0.61 & 0.68 & 0.70 & 0.65 \\
KNN & 1.46 & 1.25 & 1.95 & 1.23 & 1.47 \\
Random Forest & 0.78 & 0.86 & 0.95 & 0.83 & 0.86 \\
Torus  & 0.96 & 0.97 & 1.05 & 1.10 & 1.02 \\
ANN & 0.66 & 0.80 & \textbf{0.57} & 0.74 & 0.69 \\
\addlinespace
Simple average  & 0.60 & 0.62 & 0.69 & 0.66 & 0.64 \\
Subset average & 0.56 & 0.56 & 0.58 & \textbf{0.61} & \textbf{0.58} \\
Weighted average & \textbf{0.55} & \textbf{0.55} & 0.65 & 0.70 & 0.61 \\
SVR aggregation & 0.57 & 0.79 & \textbf{0.57} & 0.81 & 0.68 \\
\bottomrule
\end{tabular}
\end{adjustbox}
\caption{Forecasted Industrial Gas Demand: out-of-sample MAEs. Each year's best performers are in boldface.}
\label{igd}
\end{table}

\begin{table}[H]
\centering
\begin{adjustbox}{max width=\textwidth}
\begin{tabular}{@{}lccccc@{}} 
\toprule
Model  &  2015 &  2016 & 2017 & 2018 & Average  \\
\midrule
Ridge  & 3.73 & 4.15 & 4.26 & 4.48 & 4.15 \\
Lasso & 3.73 & 4.15 & 4.26 & 4.49 & 4.16 \\
Elastic Net  & 3.73 & 4.15 & 4.26 & 4.49 & 4.16 \\
SVR  & 3.41 & 3.64 & 4.33 & 4.33 & 3.93 \\
GP & 3.49 & 3.70 & 4.39 & 4.34 & 3.98 \\
KNN & 6.13 & 5.22 & 5.83 & 5.54 & 5.68 \\
Random Forest & 4.66 & 4.43 & 4.87 & 4.84 & 4.70 \\
Torus  & 3.98 & 4.48 & 4.96 & 4.94 & 4.59 \\
ANN & 3.40 & 3.97 & 4.32 & 4.41 & 4.03 \\
\addlinespace
Simple average  & 3.50 & 3.75 & 4.21 & 4.36 & 3.96 \\
Subset average & \textbf{3.26} & 3.65 & \textbf{4.17} & \textbf{4.26} & \textbf{3.84} \\
Weighted average & 3.35 & 3.71 & 4.31 & 4.31 & 3.92 \\
SVR aggregation & 3.38 & \textbf{3.62} & 4.28 & 4.37 & 3.91 \\
\bottomrule
\end{tabular}
\end{adjustbox}
\caption{Forecasted Thermoelectric Gas Demand: out-of-sample MAEs. Each year's best performers are in boldface.}
\label{tgd}
\end{table}

\begin{table}[H]
\centering
\begin{adjustbox}{max width=\textwidth}
\begin{tabular}{@{}lccccc@{}} 
\toprule
Model  &  2015 &  2016 & 2017 & 2018 & Average  \\
\midrule
Ridge               & 6.32 & 6.34 & 5.80 & 6.57 & 6.26    \\
Lasso               & 6.32 & 6.34 & 5.81 & 6.57 & 6.26    \\
Elastic Net         & 6.32 & 6.35 & 5.81 & 6.57 & 6.26    \\
SVR                 & 5.23 & 5.05 & 5.55 & 5.85 & 5.42    \\
GP                  & 5.33 & 5.23 & 5.88 & 5.82 & 5.57    \\
KNN                 & 9.04 & 9.31 & 9.97 & 9.83 & 9.54    \\
Random Forest       & 6.58 & 6.45 & 7.15 & 7.11 & 6.82    \\
Torus               & 6.56 & 6.47 & 6.40 & 7.00 & 6.61    \\
ANN                 & 5.43 & 5.50 & 5.47 & 6.08 & 5.62    \\
\addlinespace
Simple average      & 5.53 & 5.40 & 5.56 & 5.98 & 5.61    \\
Subset average      & \textbf{5.02} & \textbf{4.80} & \textbf{5.23} & \textbf{5.46} & \textbf{5.13}    \\
Weighted average    & 5.27 & 5.01 & 5.34 & 5.55 & 5.29    \\
SVR aggregation     & 5.19 & 4.91 & 5.29 & 5.79 & 5.30    \\
\addlinespace
SNAM forecast       & n.a. & n.a. & 9.62 & 8.30 & n.a.    \\
\bottomrule
\end{tabular}
\end{adjustbox}
\caption{Forecasted Italian Gas Demand: out-of-sample MAEs. Each year's best performers are in boldface.}
\label{gd}
\end{table}

\section{Conclusions} \label{conclusion}
We analyzed the industrial and thermoelectric components of Italian daily gas demand, completing a previous study concerning residential demand. We found that industrial and thermoelectric demand show different relationships with temperature and  crafted features to properly take them into account.

Several forecasting models were investigated and compared: nine base models plus four ensemble models. Aggregated models were found to be consistently more effective than base ones. In particular, in 2017 and 2018 the best ensemble model, i.e. subset average, outperformed forecasts provided by the Italian TSO.

\bibliographystyle{unsrt}
\bibliography{references}

\end{document}